\documentclass[10pt]{article} % For LaTeX2e
\usepackage[preprint]{tmlr}
\usepackage{amsmath,amsfonts,bm}

\def\eqref#1{equation~\ref{#1}}
\def\1{\bm{1}}

\DeclareMathAlphabet{\mathsfit}{\encodingdefault}{\sfdefault}{m}{sl}
\SetMathAlphabet{\mathsfit}{bold}{\encodingdefault}{\sfdefault}{bx}{n}

\usepackage{graphicx} % more modern
\usepackage{subcaption}
\usepackage{wrapfig}
\usepackage{mathrsfs}

\usepackage[document,bottom]{ragged2e}

\usepackage{url}

\usepackage[
    colorlinks=true,       % Enable colored links
    linkcolor=blue,        % Color for internal links (e.g., references within the document)
    citecolor=black,       % Color for citations
    filecolor=magenta,     % Color for file links
    urlcolor=cyan          % Color for external URLs
]{hyperref}

\usepackage{xcolor}

\definecolor{tab:blue}{RGB}{31, 119, 180}    % Matplotlib's tab:blue
\definecolor{tab:orange}{RGB}{255, 127, 14}    % Matplotlib's tab:orange
\definecolor{tab:green}{RGB}{44, 160, 44}     % Matplotlib's tab:green
\definecolor{tab:red}{RGB}{214, 39, 40}      % Matplotlib's tab:red
\definecolor{tab:purple}{RGB}{148, 103, 189}   % Matplotlib's tab:purple
\definecolor{tab:brown}{RGB}{140, 86, 75}     % Matplotlib's tab:brown
\definecolor{tab:pink}{RGB}{227, 119, 194}    % Matplotlib's tab:pink
\definecolor{tab:gray}{RGB}{127, 127, 127}    % Matplotlib's tab:gray
\definecolor{tab:olive}{RGB}{188, 189, 34}    % Matplotlib's tab:olive
\definecolor{tab:cyan}{RGB}{23, 190, 207}     % Matplotlib's tab:cyan

\definecolor{tabred}{HTML}{e74c3c} 
\definecolor{tabgreen}{HTML}{2ecc71} 
\definecolor{tabblue}{HTML}{3498db}

\title{
Rethinking Conventional Wisdom in Machine Learning: 
\\
From Generalization to Scaling
\\
}

\author{\name Lechao Xiao \email xlc@google.com \\
      \addr Google DeepMind\\
      }

\def\month{MM}  % Insert correct month for camera-ready version
\def\year{YYYY} % Insert correct year for camera-ready version
\def\openreview{\url{https://openreview.net/forum?id=XXXX}} % Insert correct link to OpenReview for camera-ready version

\begin{document}

\maketitle

\begin{abstract}
The remarkable success of large language pretraining and the discovery of scaling laws signify a paradigm shift in machine learning. Notably, the primary objective has evolved from minimizing generalization error to reducing approximation error, and the most effective strategy has transitioned from regularization (in a broad sense) to scaling up models. This raises a critical question: 
\begin{center}
\textit{Do the established principles that proved successful in the generalization-centric era remain valid in this new era of scaling?}
\end{center}

This paper examines several influential regularization-based principles that may no longer hold true in the scaling-centric, large language model (LLM) era. These principles include explicit L2 regularization and implicit regularization through small batch sizes and large learning rates. Additionally, we identify a new phenomenon termed ``scaling law crossover,'' where two scaling curves intersect at a certain scale, implying that methods effective at smaller scales may not generalize to larger ones.
Together, these observations highlight two fundamental questions within this new paradigm:
\begin{itemize}
    \item {\bf Guiding Principles for Scaling:} 
{ If regularization is no longer the primary guiding principle for model design, what new principles are emerging to guide scaling?} 
\item 
{\bf Model Comparison at Scale:} 
{ How to reliably and effectively compare models at the scale where only a single experiment is feasible?}
\end{itemize}
\end{abstract}

\begin{figure}[h!]
    \centering
        \includegraphics[width=1.\textwidth]{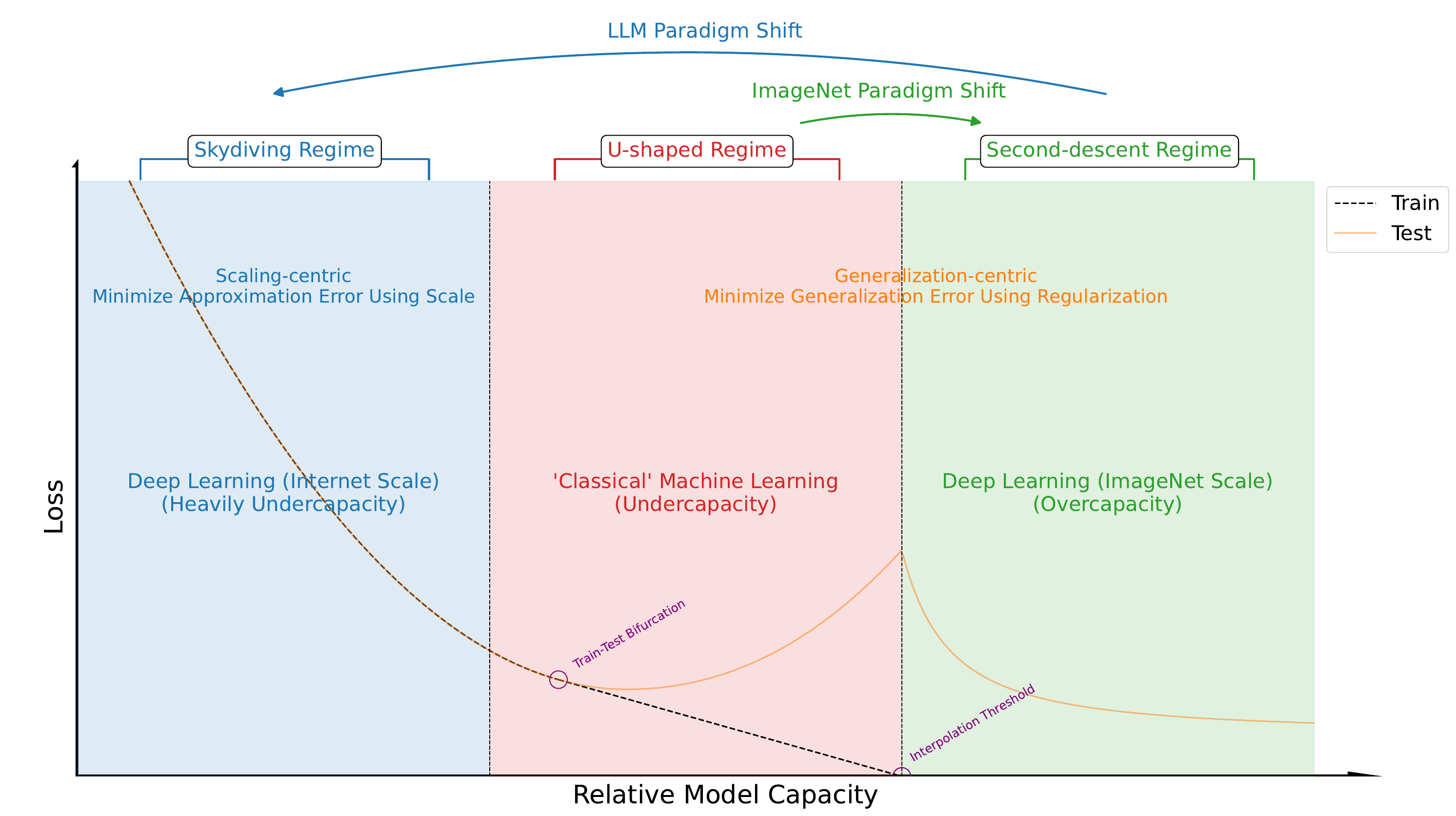}
    \caption{A proposal to reconcile  {\color{tab:red} ``Classical'' Machine Learning (U-shape)},  {\color{tab:green}  ImageNet-scale Deep Learning (Second-descent)} and 
    {\color{tab:blue} 
    Internet-scale Deep Learning (Skydiving).}}
    \label{fig:sky-diving}
\end{figure}

% The Shifting Landscape of Machine Learning
\section{Introduction}

The advent of large language pretraining \citep{devlin2018bert, radford2019language, raffel2020exploring, brown2020language}
and the emergence of scaling laws \citep{kaplan2020scaling, hoffmann2022training, achiam2023gpt, hestness2017deep} have led to a new paradigm within machine learning. This shift redirects both the primary objective and primary approach from optimizing {\it generalization} \citep{zhang2021understanding} on a fixed small dataset using {\it regularization} to reducing {\it approximation} error on a huge text corpus using {\it scale}.

More precisely, in the previous paradigm, we have far more compute than needed to interpolate the training set. To improve the performance of the model on unseen data, we need to reduce the degree of overfitting via all kinds of regularization, including inductive biases, explicit regularization (e.g., L2 Regularization), implicit regularization (e.g., large learning rate, small batch size), among many others. By contrast, in the new paradigm, we are compute constraint, i.e., we have far more data and our compute is not enough to fit such data perfectly. Existing work shows, empirically, the better the model memorizes the data (smaller approximation error), the more powerful the (foundational) model is. As such, the most effective approach to improve performance (of downstream tasks) is to throw more compute to consume the data, aka scaling up our models \citep{srivastava2023beyond}.

To distinguish these two paradigms, we refer to the former as ``generalization-centric paradigm'' and the latter as ``scaling-centric paradigm.''  We summarize several key differences below and in Fig.~\ref{fig:sky-diving}. 

\begin{enumerate}
\item 
{\bf Objective}: minimizing generalizaion error vs. minimizing approximation error. 

\item {\bf Approach}: regularization vs. scaling.

\end{enumerate}

Given these fundamental differences, blindly applying best practices from the generalization-centric paradigm to the scaling-centric paradigm may be detrimental. This note revisits several influential ideas that likely originated from the need for regularization aimed at reducing overfitting: 
\begin{itemize}
    \item Larger learning rates near the maximum stable learning rate \citep{lewkowycz2020large, li2019towards} usually generalize better. 
    \item Small batch sizes \citep{keskar2016large, smith2017bayesian} usually generalize better. 
\end{itemize}
Our empirical evidence suggests that several pieces of conventional wisdom relating to regularization in machine learning may not hold true in the scaling-centric paradigm. This raises a crucial question:

\begin{center}
\textit{``If regularization (reducing overfitting) is no longer the primary guiding principle for designing ML models, what new principles are emerging to guide scaling?''}
\end{center}

Another key distinction between the new paradigm and the old one is the immense scale involved \cite{openai2023gpt, anil2023gemini, zhang2024llasa}, which presents significant challenges both theoretically and practically. Notably, we observe a phenomenon termed ``scaling law crossover'': techniques that enhance performance at smaller scales may not translate effectively to larger ones, i.e., these techniques "overfit" to small scales. We illustrate this phenomenon with three examples. This raises a fundamental question in machine learning:

\begin{center}
\textit{``Given the potential for scaling law crossovers, how can we effectively compare models at a scale where only a single experiment is feasible?''}
\end{center}

Consequently, this new paradigm necessitates the development of novel ideas and (meta-)methodologies to understand and improve scaling.

\section{Background: Two Paradigms in Machine Learning}

% \section{Introduction} % Added a section title for clarity
The central goal in machine learning is to learn a function capable of making predictions on unseen data by understanding the underlying structure of the data.
Formally, given a training set $\mathcal{T} = \{(\mathbf{x}_i, y_i)_{1 \leq i \leq |\mathcal{T}|}\}$ drawn from some distribution $(\mathbf{x}, y) \sim \mathcal{D}$, we aim to learn a function, parameterized by $\theta \in \Omega$, that minimizes the test loss\footnote{For simplicity, we assume the irreducible loss is zero.} on unseen data:
\begin{equation}
\mathscr{L}(f_\theta; \mathcal{D}) = \mathbb{E}_{(\mathbf{x}, y) \sim \mathcal{D}} [\ell(f_\theta(\mathbf{x}), y)]
\end{equation}

In practice, we learn $f_\theta$ by minimizing the training loss (the average loss on the training set): 
%using gradient-based optimization methods:

\begin{equation}
\mathscr{L}(f_\theta; \mathcal{T}) := \frac{1}{|\mathcal{T}|} \sum_{(\mathbf{x}, y) \in \mathcal{T}} \ell(f_\theta(\mathbf{x}), y) 
\end{equation}
This approach is known as empirical risk minimization (ERM). The test error can be decomposed into the sum of generalization gap and training error (approximation error) 

\begin{equation}
 \underbrace{\mathscr{L}(f_\theta; \mathcal{D})}_{\text{Test Error}}\quad  =\quad \underbrace{\left(
\mathscr{L}(f_\theta; \mathcal{D})  - \mathscr{L}(f_\theta; \mathcal{T}) 
\right)}_{\text{Generalization Gap}} \quad\quad
+ \underbrace{\mathscr{L}(f_\theta; \mathcal{T}) }_{\text{Approximation Error}}
\end{equation}

\begin{wrapfigure}{r}{0.45\textwidth}
  \begin{center}
    \includegraphics[width=0.4\textwidth]{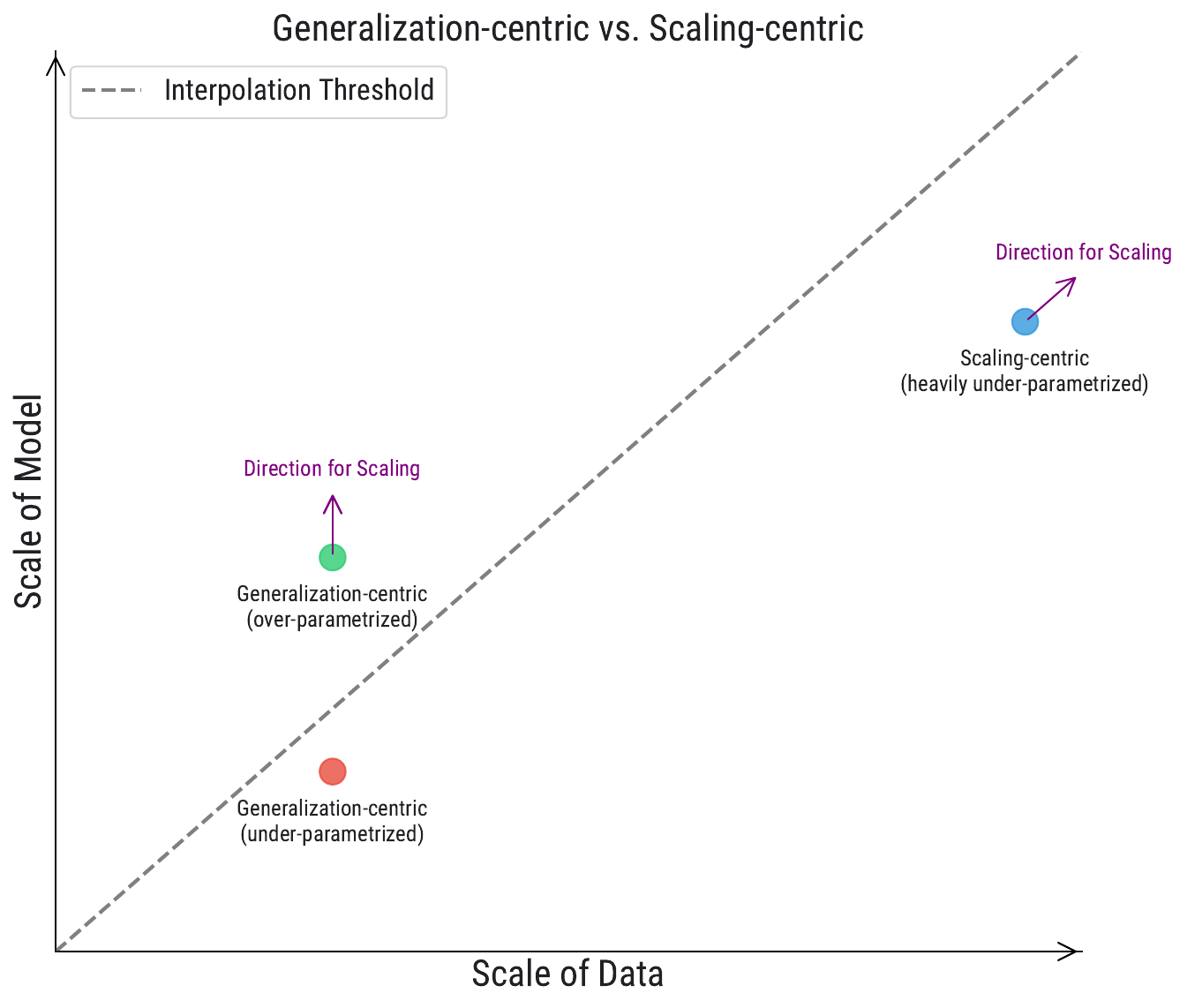}
  \end{center}
  \caption{Generalization vs Scaling Paradigms.}
  \label{fig:data-vs-model}
\end{wrapfigure}

% This approach is known as empirical risk minimization (ERM). 

We discuss two paradigms in machine learning, distinguished by the relative and absolute scales\footnote{We use the term ``scale'' broadly in this context to encompass both the capacity of the model and the complexity of the data.} of the data and model:

\begin{itemize}

\item \textbf{Generalization-centric paradigm:}

Data scale is relatively small. This paradigm further divides into two sub-paradigms:
    \begin{itemize}
    \item \textbf{Classical bias-variance trade-off (U-shaped) regime:} Model capacity is intentionally constrained below the interpolation threshold (red dot \textcolor{tabred}{$\bullet$} in Fig.~\ref{fig:data-vs-model}).
    In this regime, both {\it Generalization Gap} and {\it Approximation Error} are non-negligible.  
    \item \textbf{Modern over-parameterized (second-descent) regime:} Model scale significantly surpasses data scale (green dot \textcolor{tabgreen}{$\bullet$} in Fig.~\ref{fig:data-vs-model}).
    In this regime, {\it Approximation Error} is negligible. 
    \end{itemize}

\item \textbf{Scaling-centric paradigm:} Large data and model scales, with data scale exceeding model scale (blue dot \textcolor{tabblue}{$\bullet$} in Fig.~\ref{fig:data-vs-model}).
In this regime, the {\it  Generalization Gap} is negligible. 
\end{itemize}

\subsection{The Bias-Variance Trade-off and the U-shape Regime}
\begin{figure}[t!]
  \centering
  \begin{tabular}{cc}
  \includegraphics[height=0.15\textheight]{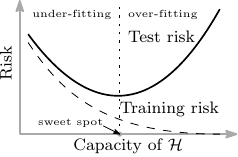} &
  \includegraphics[height=0.15\textheight]{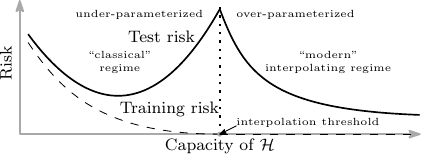}     \\
  {\bf (a)} & {\bf (b)}
  \end{tabular}
  \caption{{\bf U-Shaped (a) and double-descent curves (b). Figure is from \cite{belkin2019reconciling}}} 
    % ({\bf a}) The classical \emph{U-shaped risk curve} arising from the bias-variance trade-off.
    % ({\bf b}) The \emph{double descent risk curve}, which incorporates the U-shaped risk curve (i.e., the ``classical'' regime) together with the observed behavior from using high capacity function classes (i.e., the ``modern'' interpolating regime), separated by the interpolation threshold.
    % The predictors to the right of the interpolation threshold have zero training risk.
    % }
  \label{fig:double-descent}
\end{figure}

% \vspace{-0.5cm}
% \begin{wrapfigure}{r}{0.45\textwidth}
%   \begin{center}
%     \includegraphics[width=0.4\textwidth]{figures/july5/data-model-scales (5).pdf}
%   \end{center}
%   \caption{Generalization vs Scaling Paradigms.}
%   \label{fig:data-vs-model}
% \end{wrapfigure}
In this classical setting, the complexity of the training set $\mathcal{T}$ is typically smaller than the richness of the function class $\mathcal{H} = \{f_\theta: \theta \in \Omega\}$
and the absolute scales of the data and models are small. Conventional wisdom in machine learning suggests that one needs to carefully control the complexity of the function space $\mathcal{H}$ \citep{belkin2019reconciling} to balance  {\it Generalization Gap} and {\it Approximation Error}:
\begin{enumerate}
    \item If $\mathcal{H}$ is too small (underfitting), all functions in $\mathcal{H}$ will have high bias, i.e. high {\it Approximation Error}. This leads to a large training error,  and thus a large test error.
    \item If $\mathcal{H}$ is too large (overfitting), the learned function may overfit the training data, leading to high variance. This results in a small training error but a large {\it Generalization Gap} (the difference between test and training errors), and thus a large test error. 
\end{enumerate}
This observation is known as the bias-variance trade-off, a fundamental result in machine learning \citep{hastie2009elements}. It traditionally suggests a U-shaped error curve (Figure~\ref{fig:double-descent} (a)), with optimal test error achieved by balancing bias and variance \citep{hastie2009elements}. It was believed that {test loss} would increase monotonically after this optimal trade-off point due to overfitting, where models capture noise in the training data. As such, the optimal function class was thought to be in the undercapacity/under-parameterized regime, in which the functions cannot perfectly fit (interpolate) the training data. Regularization methods \citep{hastie2009elements} like L2,  weight decay, Lasso, Early Stopping etc. are used to achieve the optimal trade-off.

\subsection{Over-parameterization and the Second-descent Regime}

The success of deep neural networks in tasks like image recognition around 2012 \citep{krizhevsky2012imagenet} marked a (sub-)paradigm shift inside generalization-centric machine learning.  Over-parameterized neural networks, possessing more parameters than required to perfectly fit the training data (interpolation threshold), surprisingly continued to improve as they became even more over-parameterized, surpassing the performance of under-parameterized models \citep{he2016deep, neyshabur2018towards}. This phenomenon, where increasing model complexity beyond the interpolation threshold leads to improved performance, offered a new perspective beyond the classical bias-variance trade-off. It inspired the proposal of the ``double-descent curve'' (Figure~\ref{fig:double-descent} (b)) to accommodate both the traditional U-shaped curve in the under-parameterized regime and the observed single-descent curve in the over-parameterized regime \citep{belkin2019reconciling}.

Key questions arose: why do over-parameterized neural networks generalize well, exhibiting benign overfitting \citep{bartlett2020benign}, and how can we design algorithms and architectures to improve generalization? A central question is why some global minima generalize much better than others in over-parameterized models \citep{zhang2021understanding, neyshabur2018towards}. This led to research on implicit regularization \citep{neyshabur2017implicit} in neural networks, aiming to provide theoretical and practical insights.
Several key findings emerged:

\paragraph{Large Learning Rate is Better.} Practitioners observed better generalization with near-maximal stable learning rates. Explanations include the flat minima hypothesis (SGD noise helps escape sharp minima \citep{keskar2016large}), the edge-of-stability concept (larger learning rates decrease hessian curvature) \citep{cohen2020gradient}, and escaping the linearization regime (large learning rates help networks learn useful representations) \citep{lewkowycz2020large}.

\paragraph{Small Batch Size is Better \citep{keskar2016large, smith2017bayesian}.} A critical batch size exists, below which generalization does not worsen with decreasing batch size. Above this, performance degrades, resembling a ReLU-shaped curve. This is attributed to gradient noise from small batches biasing towards flatter minima, while large batches lead to sharp minima.

These insights were developed to improve and explain generalization through implicit or explicit regularization, which is believed to reduce variance, guiding the network to find better minima  \citep{foret2020sharpness}. 

Fig.~\ref{fig:typical-dynamics} (a) depicts typical learning dynamics in this paradigm: training and test error curves diverge after a period of training, creating a generalization gap, while training error (classification) reaches a global minimum. Minimizing this generalization gap remains a central focus in this paradigm

We summarize two key principles in the generalization-centric paradigm:

\begin{itemize}
    \item \textbf{Guiding Principles for Generalization}: Overfitting is a core challenge, and ``regularization'' serves as the primary guiding principle for understanding and improving generalization.
    \item \textbf{Model Comparison via Validation Set}: Due to the relatively smaller scale of problems in this paradigm, we can afford to train multiple models and rely on hold-out validation sets for model comparison, which is a reliable and effective approach.
\end{itemize}

However, in the scaling paradigm, we may lose the advantages offered by both of these principles.

\subsection{Heavy Under-parameterization and the Skydiving Regime}
Breakthroughs in large language model pretraining leads us to a scaling-centric paradigm, distinguished from the previous generalization-centric paradigm by two key features. 
 First, the complexity of training data $\mathcal{T}$ far surpasses the capacity of the models \citep{raffel2020exploring, brown2020language, hoffmann2022training, touvron2023llama}
and the training loss remains far from reaching a plateau. Second, both the data and the models themselves operate at a scale vastly larger than in previous paradigms, as illustrated in Figure~\ref{fig:data-vs-model} blue dot \textcolor{tabblue}{$\bullet$}.

\begin{figure}[t!]
  \centering
  \begin{tabular}{cc}
  \includegraphics[width=0.45\textwidth]{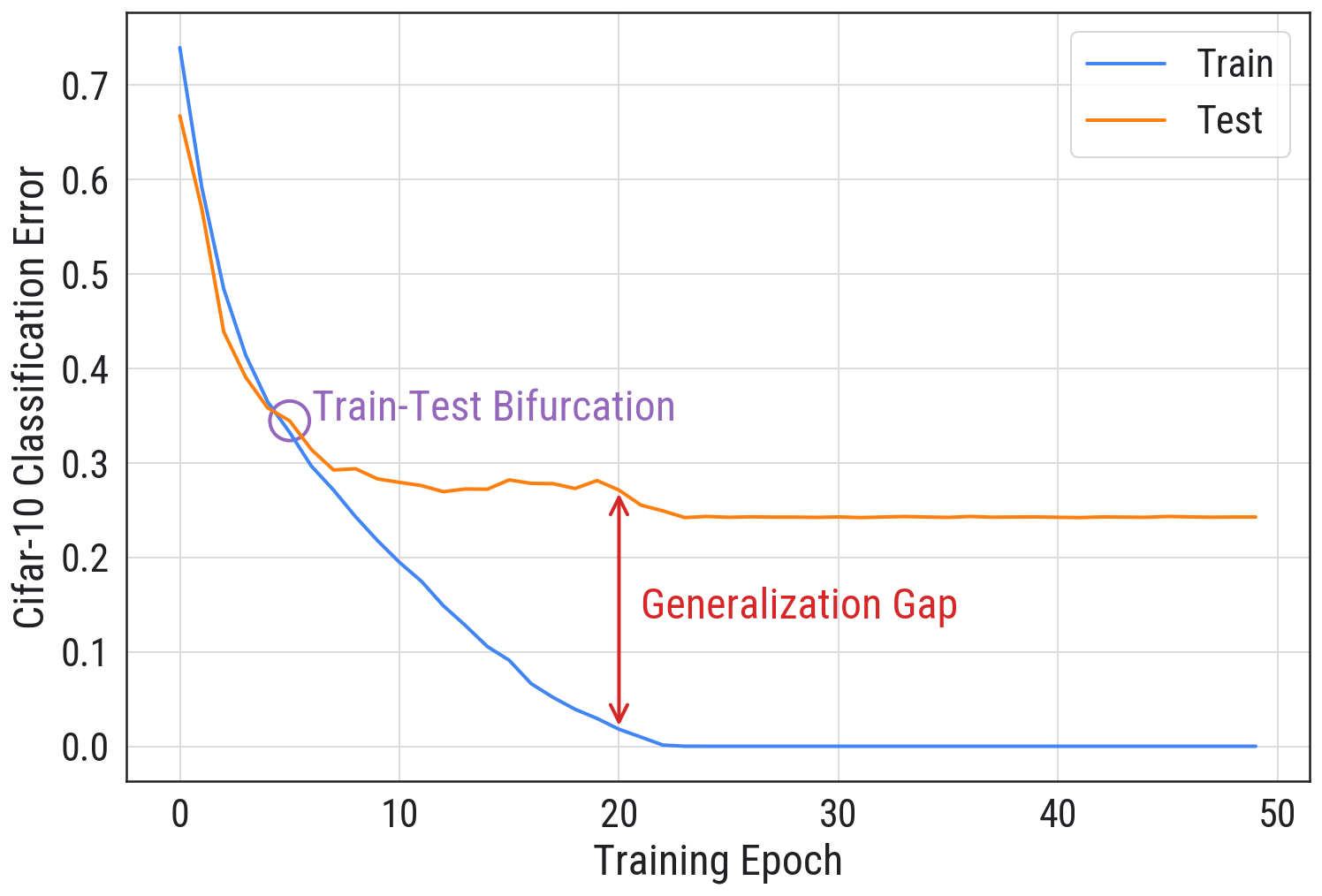} &
  \includegraphics[width=0.47\textwidth]{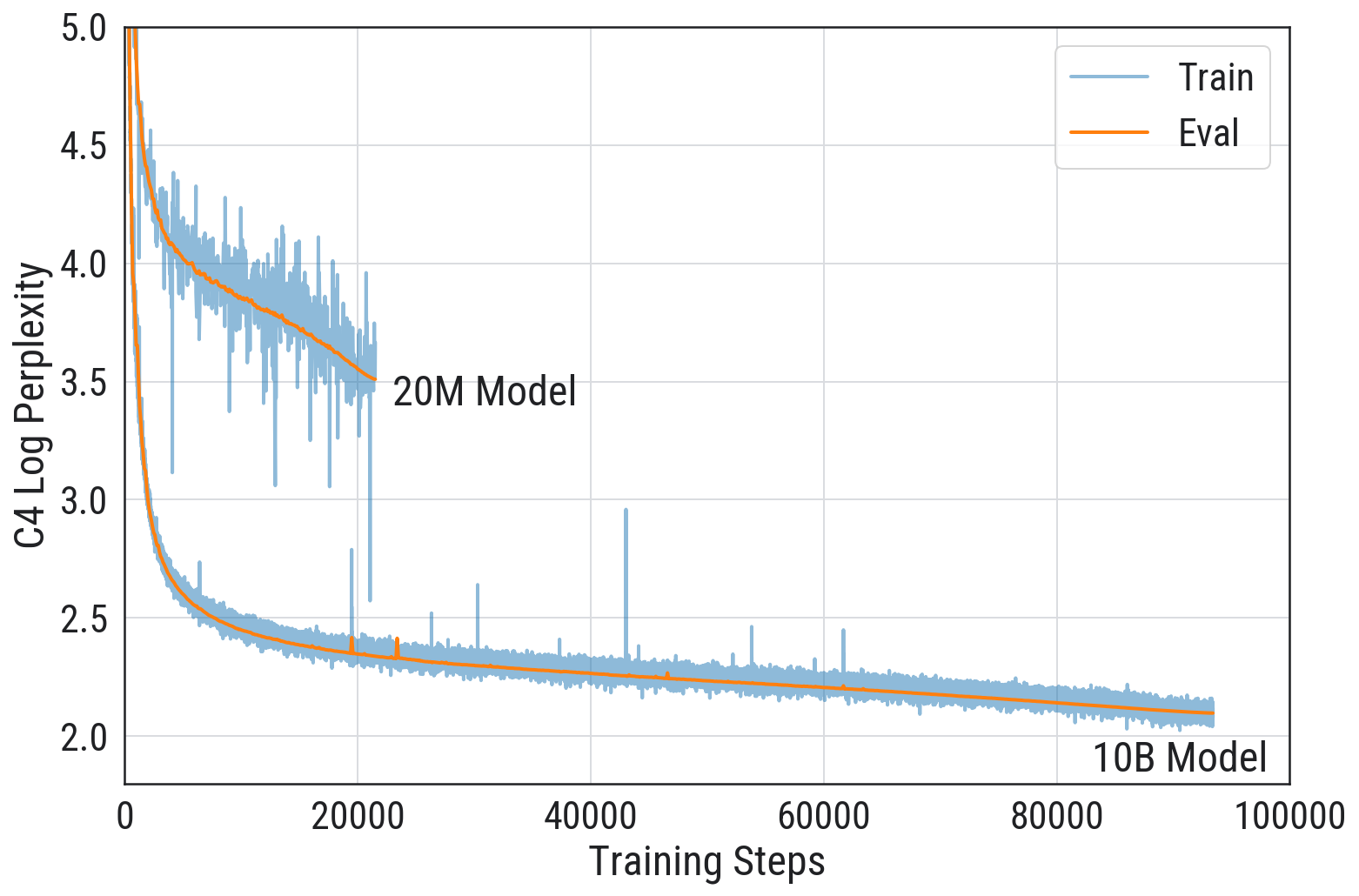}\\
  {\bf (a) Image Classification (Cifar-10)} & {\bf (b) Language Modeling (C4)}
  \end{tabular}
  \caption{
  {\bf Learning Dynamics: Generalization (Image Classification) vs. Scaling (Language Model Pretraining).}
  {\bf (a)}
  ResNet-18 on CIFAR-10. Training and test error curves initially overlap, then diverge, forming a generalization gap. Minimizing this gap is the central objective as the network easily interpolates training data.
  {\bf (b)}.
 Decoder-only transformer on C4. Evaluation curves consistently remain within training curves throughout training, even when the model size and compute is scaled up by a factor of 500 and 250,000, respectively.
  } 
    % ({\bf a}) The classical \emph{U-shaped risk curve} arising from the bias-variance trade-off.
    % ({\bf b}) The \emph{double descent risk curve}, which incorporates the U-shaped risk curve (i.e., the ``classical'' regime) together with the observed behavior from using high capacity function classes (i.e., the ``modern'' interpolating regime), separated by the interpolation threshold.
    % The predictors to the right of the interpolation threshold have zero training risk.
    % }
  \label{fig:typical-dynamics}
\end{figure}

Figure \ref{fig:typical-dynamics} (b) illustrates the typical learning dynamics in this paradigm: test and training error curves remain closely aligned throughout training, even when model size and compute are scaled up by factors of 500 and 250,000, respectively. The training error has not yet reached its global minimum, suggesting further scaling up either or both the model size and dataset size could lead to improved performance.  This behavior corresponds to the ``skydiving (blue)'' regime depicted in Figure \ref{fig:sky-diving}, preceding the U-shaped bias-variance trade-off curve (Fig.~\ref{fig:double-descent} (a)). In this regime, training and test errors on a holdout evaluation set are nearly identical. Consequently, the primary goal shifts from mitigating overfitting to minimizing the pretraining (approximation) error, as the generalization gap has yet to emerge.

This divergence in primary objectives suggests that conventional wisdom from generalization-centric machine learning might not readily apply to the new paradigm. Specifically, ``regularization'' may no longer be the driving force for performance improvement (see Section \ref{sec:is-reg}) nor the guiding principle for understanding scaling. Furthermore, Section \ref{sec:no-free-lunch} explores the challenges arising from the immense scale involved. In particular, the hold-out validation set method may not be applicable for model comparison in this setting.

\section{Architecture and Optimizer}\label{sec:architetures and data}
We use decoder-only transformer architecture \citep{vaswani2017attention}. All language models are trained on the C4 dataset \citep{raffel2020exploring}. 
We use the open-source NanoDO codebase \citep{nanodo} for our training process. Specific architectural details are listed below.

\begin{itemize}
    \item {\bf Rotary} \citep{su2024roformer} Positional Embedding. 
    \item {\bf QK-Norm} \citep{Gilmer2023, pmlr-v202-dehghani23a}, i.e., two Layer Normalization layers are applied to the queries and keys before the dot-product attention computation
    \item {\bf Untying} the head from the embedding, i.e., we do not use weight tying of the first and last layer. 
    \item {\bf Gelu} \citep{hendrycks2016gaussian} activation with $F=4D$, where $D$ and $F$ are the model dim and hidden dim of the MLP, resp. However, we use Geglu \citep{shazeer2020glu} in some experiments. 
    \item The head dimension of query and key is set to $d_{\text{head}} = 64$, resulting in $H = D / d_{\text{head}}$ attention heads throughout this paper.
    \item The sequence length is $S=512$. 
    \item The vocabulary size is $V=32101$.
    \item The total number of parameters in the backbone is approximately $\mathscr{N} \approx 12D^2L$, where $D$ represents the model dimension and $L$ is the number of layers in the transformer. 
    \item Most models are trained to Chinchilla optimality \citep{hoffmann2022training}, utilizing a total of $\mathscr{D} = 20 \times (12D^2L + DV)$ tokens.  %$\mathscr{ND}$
    \item 
    The total compute is estimated using the ``$\mathscr{F} = 6\mathscr{ND}$'' formula \cite{kaplan2020scaling}, where the estimated number of floating-point operations (FLOPs) $\mathscr{F}$ is $6\times \text{Number of Parameters}  (\mathscr{N}) \times \text{Number of Tokens} (\mathscr{D})$.
\end{itemize}

\paragraph{Optimizer.}
Our default optimizer is AdamW \citep{kingma2014adam, loshchilov2017decoupled, deepmind2020jax} with $\beta_1=0.9$, $\beta_2=0.95$, $\epsilon=\text{1e-20}$ and coupled weight decay $\lambda =0.1$. 

\section{Is Regularization Needed?}
\label{sec:is-reg}

As discussed earlier, regularization plays a pivotal role in the generalization-centric paradigm. It effectively mitigates overfitting and bridges the gap between training and test losses. In this section, we revisit three popular regularization techniques commonly employed in machine learning: explicit L2 regularization, and the implicit regularization effects of large learning rates and small batch sizes.

While the conclusion of this section — that explicit/implicit regularization is \textit{not} necessary in the absence of overfitting — may be obvious to many, it is useful to re-examine our prior beliefs in the context of the scaling-centric paradigm.

\begin{figure}[b!]
    \centering
    \includegraphics[width=\textwidth]{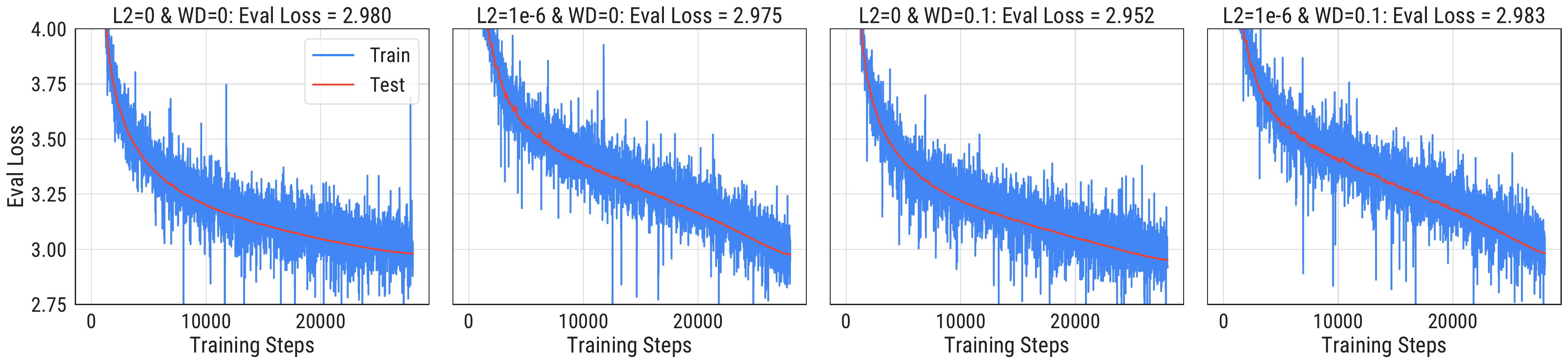}
    \caption{Training dynamics of four transformer models. From left to right: no L2 and no weight decay, small L2 and no weight decay, no L2 but with weight decay, with both L2 and weight decay. }
    \label{fig:is-l2-reg-useful}
\end{figure}

\subsection{Does L2 Regularization Improve Performance?}

To assess the impact of L2 regularization across different training regimes, we compare its benefits in scaling-centric versus generalization-centric settings.

\begin{wrapfigure}{r}{0.4\textwidth}
  \begin{center}
    \includegraphics[width=0.4\textwidth]{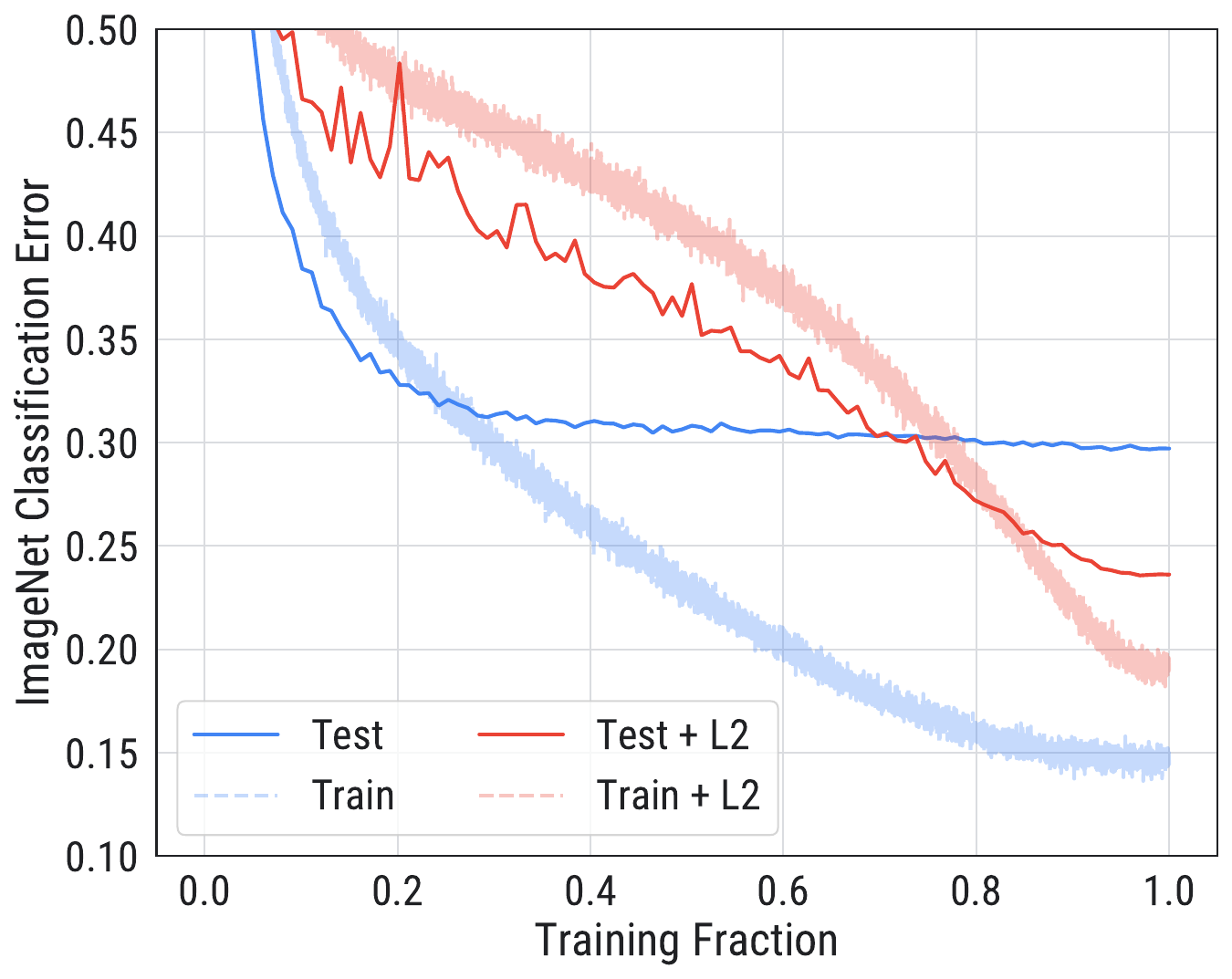}
  \end{center}
  \caption{ResNet-50 on ImageNet. L2 regularization reduces overfitting and improves generalization. }
  \label{fig: l2-reg-imagenet}
\end{wrapfigure}

First, we showcase the usefulness of L2 in improving generalization. 
We trained a ResNet-50 on ImageNet using \href{https://github.com/google/flax/tree/main/examples/imagenet}{Flax's default settings}, which include L2 regularization ($\lambda=0.0001$). We then trained a second model without L2 regularization ($\lambda=0$). As shown in Fig.~\ref{fig: l2-reg-imagenet}, L2 regularization substantially boosts test accuracy by $6\%$ (0.764 vs. 0.703). While the model without L2 achieves a higher training accuracy ($\sim 0.856$ vs. $ \sim 0.807$), this clearly highlights the effectiveness of L2 regularization in reducing overfitting and improving generalization.

Next, we provide evidence that L2 regularization may not be useful for language model pretraining. To do so, we trained four 151M transformers, varying the use of L2 regularization and weight decay. The training dynamics are presented in Figure \ref{fig:is-l2-reg-useful}, leading to the following observations:

\begin{enumerate}
    \item Operating within the ``skydiving regime,'' where training and test losses closely track each other, implying no generalization gap, the addition of L2 regularization does not alter this behavior.
    
    \item In our experiments, the baseline model without L2 or weight decay achieves a final evaluation loss of 2.980. Introducing L2 regularization alone (2.975) or in combination with weight decay (2.983) does not appear to improve test loss. However, employing weight decay independently leads to a notable performance gain (2.952).
\end{enumerate}

\paragraph{Discussion.} While L2 regularization is widely acknowledged to reduce overfitting and thus enhance generalization performance, our preliminary experiments indicate that it may not offer similar benefits for language model pretraining. This aligns with current practices, as flagship language models such as GPT-3 \citep{brown2020language}, PALM \citep{chowdhery2023palm}, Chinchilla \citep{hoffmann2022training}, Llama-2 \citep{touvron2023llama} and DeepSeek-V2 \citep{deepseekai2024deepseekv2strongeconomicalefficient} do not employ L2 regularization. While weight decay is widely used in training language models, our observations are consistent with \cite{andriushchenko2023we} in that it does not play a conventional regularizer role.  Guiding principles are needed to deepen our understanding of weight decay in this context \citep{wang2024set}.

\subsection{Does Maximal Stable Learning Rate Perform Better?}\label{sec:maximal-learning-rate}

\begin{figure}[t!]
\centering
     \begin{subfigure}[h]{0.4\textwidth}
         \centering
         \includegraphics[width=\textwidth]{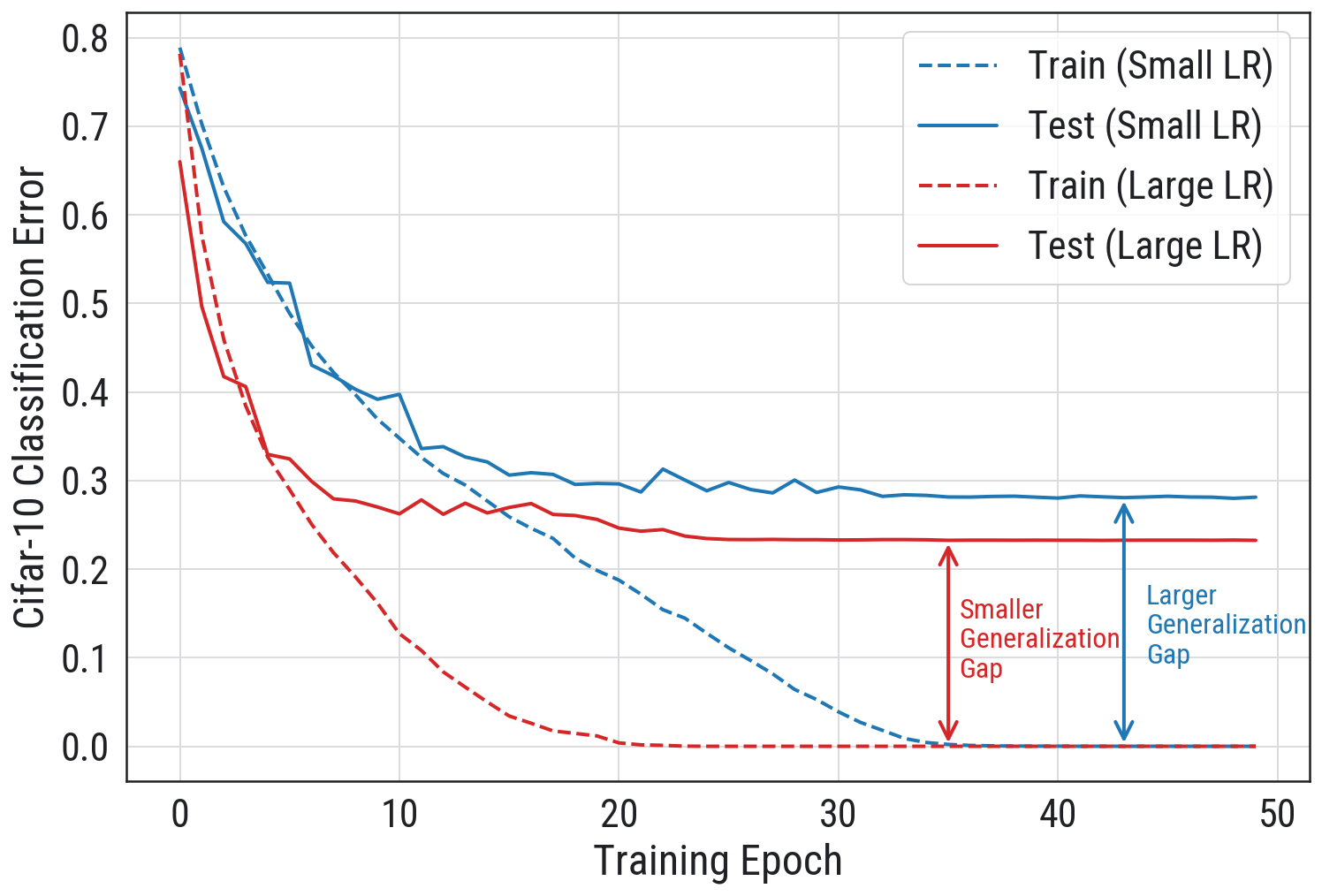}
         \caption{Large Learning Rate is Better}
         \label{fig:small-vs-large-lr-cifar10.}
     \end{subfigure}
    %   \hfill
  \centering
     \begin{subfigure}[h]{0.4\textwidth}
         \centering
         \includegraphics[width=\textwidth]{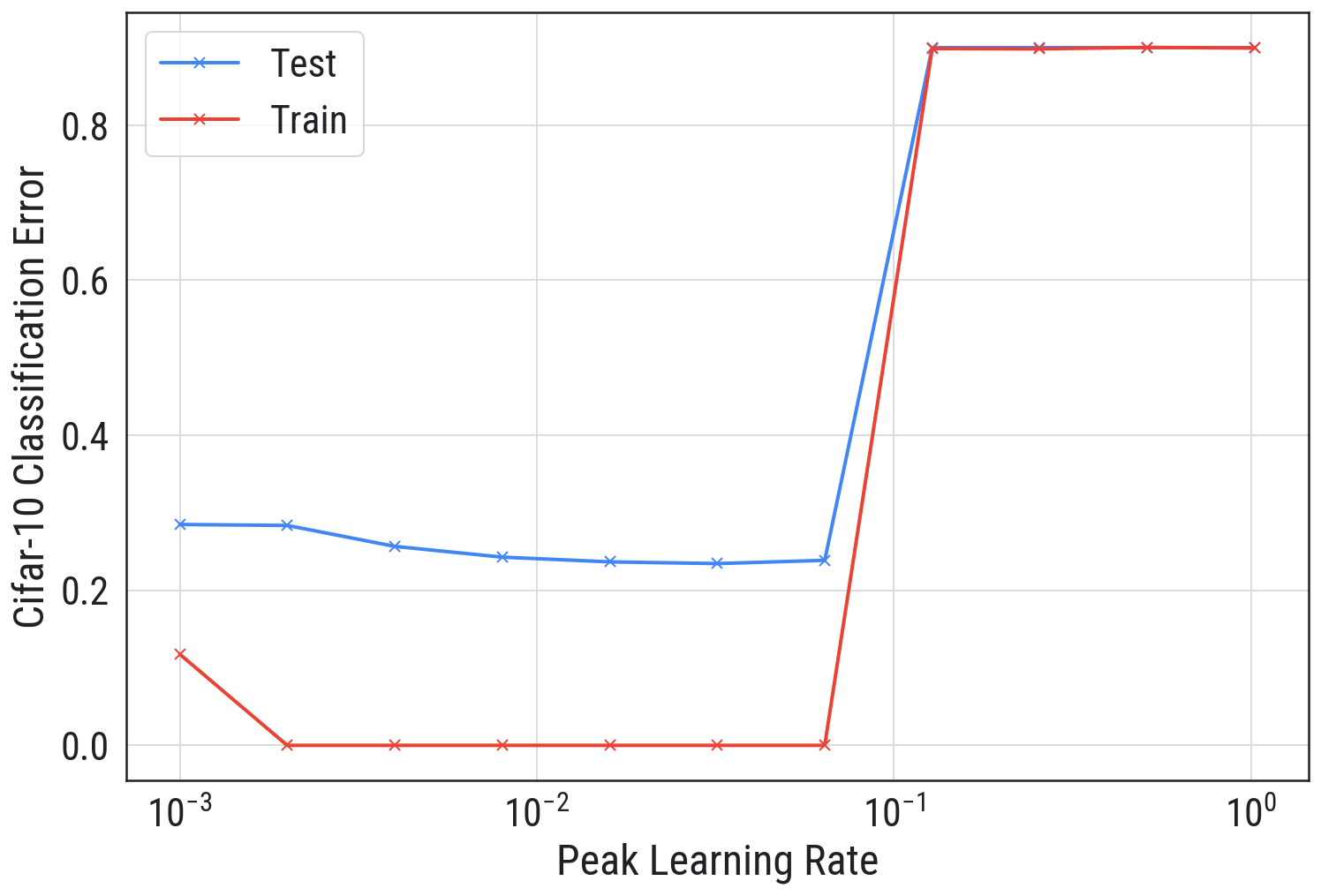}
         \caption{Optimal LR is near Max Stable LR}
         \label{fig:cifar-large-is-better.}
     \end{subfigure}
      \hfill
        \caption{{\bf Large Learning Rate Improves Generalization.} (a) Networks trained with large or small learning rates can achieve perfect training accuracy, but large learning rates generalize better. 
        (b)
        The optimal learning rate is often near the maximum stable rate.}
        \label{fig:CIFAR-10-large-lr}
\end{figure}

Conventional wisdom in neural network training often favors using a larger learning rate, possibly near the maximum stable value, as this is believed to improve generalization performance \citep{ li2019towards, lee2020finite, lewkowycz2020large}. This practice is attributed to the implicit regularization of stochastic gradient descent (SGD). While various learning rates can achieve perfect training accuracy for small datasets, e.g. CIFAR-10, the gradient noise from larger learning rates is thought to guide SGD towards better minima. This phenomenon has been explored in several influential studies, including those on flat minima \citep{hochreiter1997flat, keskar2016large, dinh2017sharp, foret2020sharpness}, the edge of stability \citep{cohen2020gradient, agarwala2022second, damian2022self, Gilmer2023},
and escaping the NTK regime \citep{jacot2018neural, chizat2019lazy, lee2019wide, lewkowycz2020large, yang2021tensor, pmlr-v125-woodworth20a, karkada2024lazy}. Although no rigorous theory fully explains why the maximal stable learning rate improves performance, it's considered by many a good practice in training neural networks. We reproduce this insight by training a ResNet-18 on CIFAR-10.  
Figure \ref{fig:small-vs-large-lr-cifar10.} shows the regularization benefits of using a larger learning rate: while both large and small learning rates lead to zero classification error, the larger learning rate results in a smaller test error.
Fig.~\ref{fig:cifar-large-is-better.} shows that the optimal learning rate is near the maximal stable learning rate. 

Does this conventional wisdom apply to training language models? Since the primary principle behind this practice is improving generalization (via implicit regularization) - different from the primary goal in the scaling-centric paradigm - we might expect it to not hold true. In the following, we provide empirical evidence supporting this hypothesis.

\begin{figure}[t!]

     \centering
     \begin{subfigure}[h]{0.4\textwidth}
         \centering
         \includegraphics[width=\textwidth]{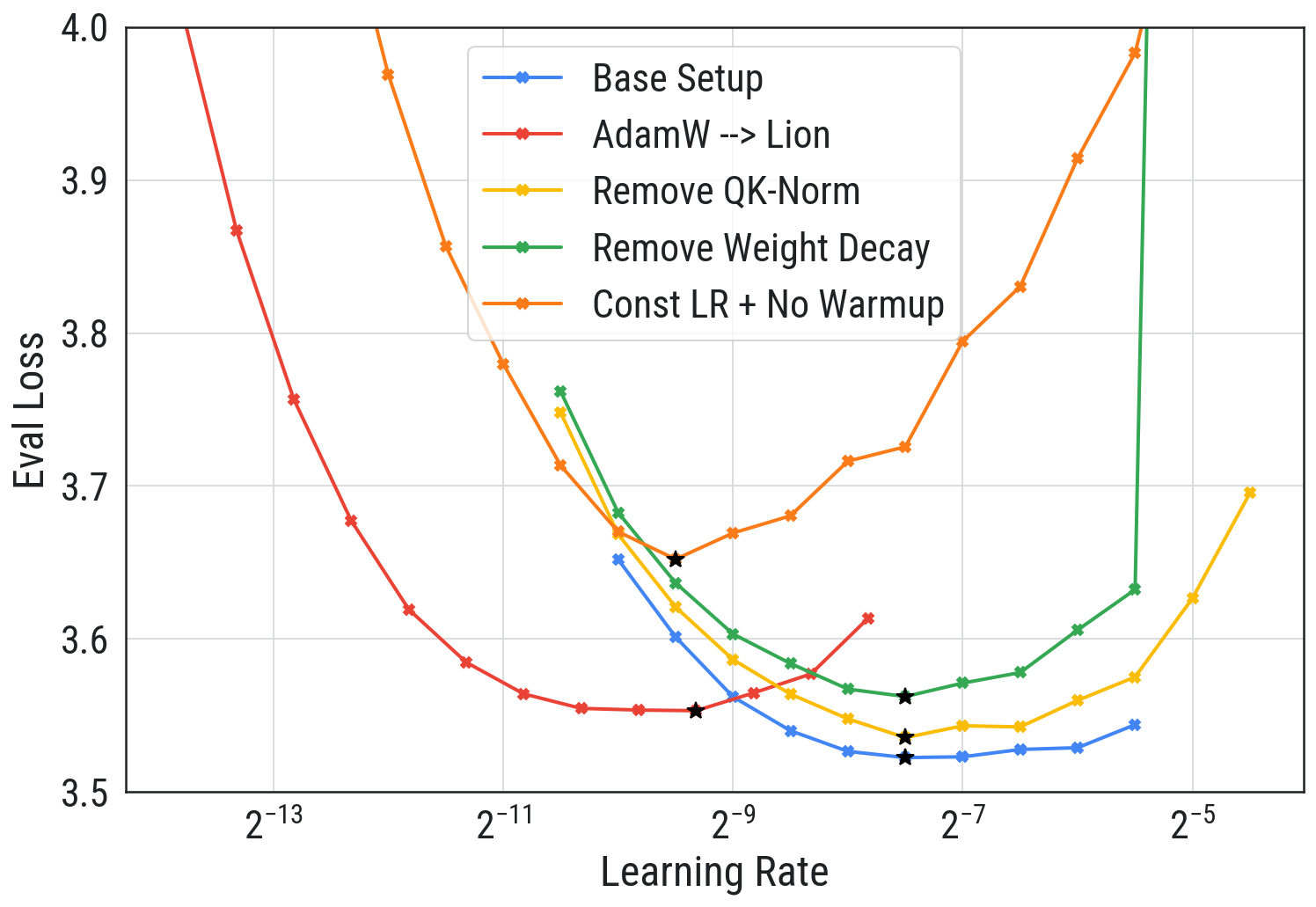}
         \caption{Learning Rate is moderate}
         \label{fig:learning rate is moder rate.}
     \end{subfigure}
    %   \hfill
     \begin{subfigure}[h]{0.4\textwidth}
         \centering
         \includegraphics[width=\textwidth]{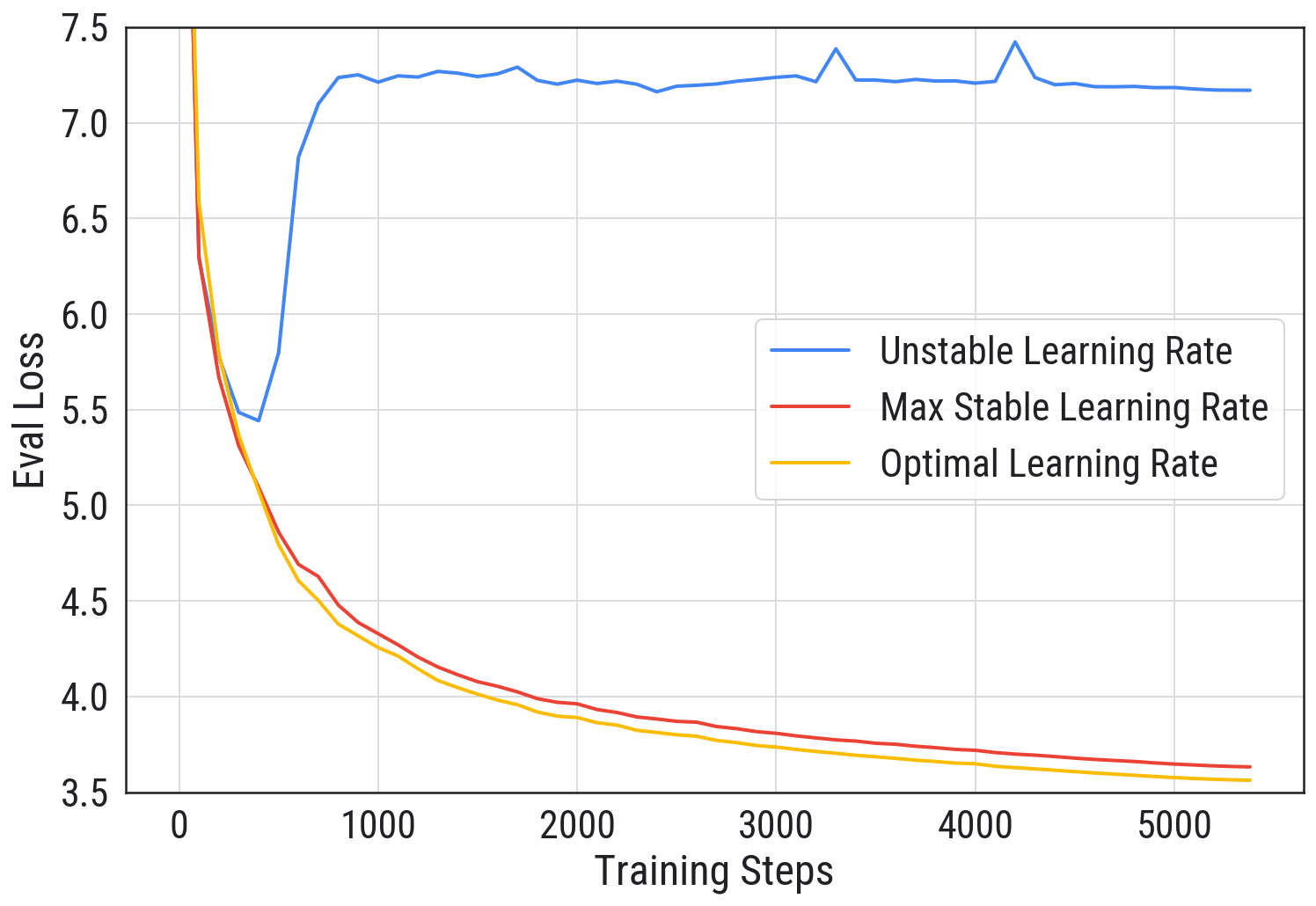}
         \caption{Model Size 151M}
         \label{fig:dynamics-large-learning-rate.}
     \end{subfigure}
     \hfill
        \caption{{\bf Optimal Learning Rates Are Significantly Lower Than Maximal Stable Learning Rates.} (a) Loss vs. learning rate curves reveal U-shaped relationships, with optimal learning rates far below stability limits. (b) Despite smooth training curves, maximal stable rates consistently underperform.
 }
        \label{fig:Learning Rates  experiments}
\end{figure}

\paragraph{Experiment Setup.} To investigate the optimal learning rate for training language models, we trained our base model to Chinchilla-Optimal while varying the learning rates. We further test the robustness of our findings by introducing several interventions:
\begin{enumerate}
    \item 
Switching from AdamW  to Lion \citep{chen2024symbolic} (Fig.~\ref{fig:learning rate is moder rate.}), 
\item Removing QK-Norm (Fig.~\ref{fig:learning rate is moder rate.}), 
\item Eliminating weight decay (Fig.~\ref{fig:learning rate is moder rate.}), 
\item Removing warmup and learning rate schedule (Fig.~\ref{fig:learning rate is moder rate.}), 
\item Adjusting batch size and changing model size (Fig.~\ref{fig:batch size vs learning rate 19m.} and Fig.~\ref{fig:batch size vs learning rate 151m.})
\end{enumerate}
\paragraph{Experimental Results.}  Across all these setups, the relationship between loss and learning rate consistently exhibited a U-shape curve.  The optimal learning rate was far from the maximal stable learning rate often favored in traditional neural network training.

\paragraph{Discussion.} Contrary to conventional wisdom, our findings reveal that the optimal learning rate for large language models is significantly lower than the maximal stable value\footnote{This observation has been recognized (implicitly or explicitly) in existing work, e.g. \citet{wortsman2023small,zhao2024deconstructing} among many others.} previously assumed. This suggests that traditional regularization-based theories may not fully explain the dynamics of optimal learning rates in the context of training language models.  Further research is needed to elucidate the complex relationship between optimal learning rate, model scale, and other training factors (see \cite{paquette2024}). %, such as Chinchilla-Optimal vs. fixed training horizons.

% \paragraph{Experiment Setup.} We trained our base model to Chinchilla-Optimal performance with various learning rates. To demonstrate the robustness of our claim, we introduced several interventions: (1) changing the batch size, (2) changing the model size, (3) changing the optimizer from AdamW to Lion, (4) removing QK-Norm, and (5) removing weight decay.

% \paragraph{Experimental Results.} In all the above setups, the loss vs. learning rate curve consistently exhibited a U-shape. The optimal learning rate was always moderate and far from the maximal stable learning rate.

% \paragraph{Conclusion.} Contrary to conventional wisdom, the optimal learning rate for large language models (LLMs) is moderate and not close to the maximal stable learning rate. Traditional regularization-based theories may not fully explain our observations regarding optimal learning rates in this context. New theories and principles are needed to understand and predict the relationship between the optimal learning rate, scale, and other training factors.

\subsection{Does Small Batch Size Perform Better?}\label{sec:small batch size}
\begin{figure}[t!]
    \centering
        \includegraphics[width=0.8\textwidth]{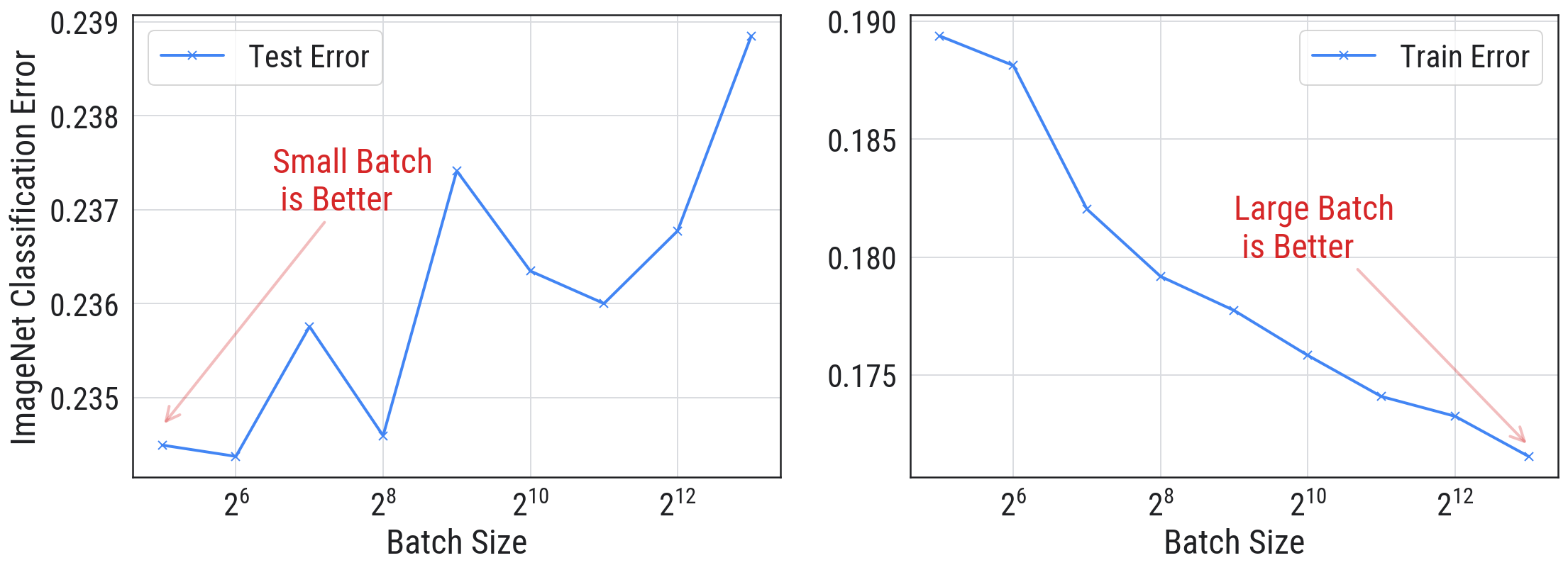}
    \caption{{\bf Small Batch Size performs better for ImageNet.}
    Small batch size achieves better {\it Test Error} (left) but worse {\it Training Error} (right), vice versa for large batch size.} 
    \label{fig:imagenet}
\end{figure}
In generalization-centric ML, a common observation is that, given the same computational budget (measured by the total number of training epochs), algorithms employing smaller batch sizes tend to generalize better than those with larger batches beyond a critical threshold \citep{keskar2016large, smith2017bayesian, shallue2019measuring, mccandlish2018empirical}. This phenomenon is often attributed to the increased gradient noise associated with smaller batches, which acts as an implicit regularizer, mitigating overfitting. It's hypothesized that the noise helps guide the optimization process towards minima that generalize better.

We replicate this conventional wisdom in Fig.~\ref{fig:imagenet} using the ImageNet example in \citep{flax2020github}. With a fixed number of training epochs (100), the trend demonstrates that smaller batch sizes lead to better test performance, albeit with worse training performance, effectively reducing overfitting.

However, does this conventional wisdom about small batch sizes translate to the realm of LLM training? Given that the primary benefit of small batches is their regularization effect, and regularization may not be the primary concern for LLMs, we have reason to question this assumption.

% Our empirical evidence suggests a different picture. We observe a U-shaped relationship between model performance and the logarithm of batch size.  This implies that both excessively small and excessively large batch sizes can be detrimental to performance. 
% Furthermore, we find that the optimal batch size increases as the model scale grows.
\paragraph{Experiment Setup.} We trained two models (19M and 151M parameters) to Chinchilla-Optimal performance, varying batch sizes in $\{16, 32, 64, 128, 256, 512, 1024\}$. For each batch size, an 11-point grid search was performed to identify the optimal learning rate. To assess variability, the 19M model was trained with five different random seeds.

\begin{figure}[b!]
     \centering
     \begin{subfigure}[h]{0.3\textwidth}
         \centering
         \includegraphics[width=\textwidth]{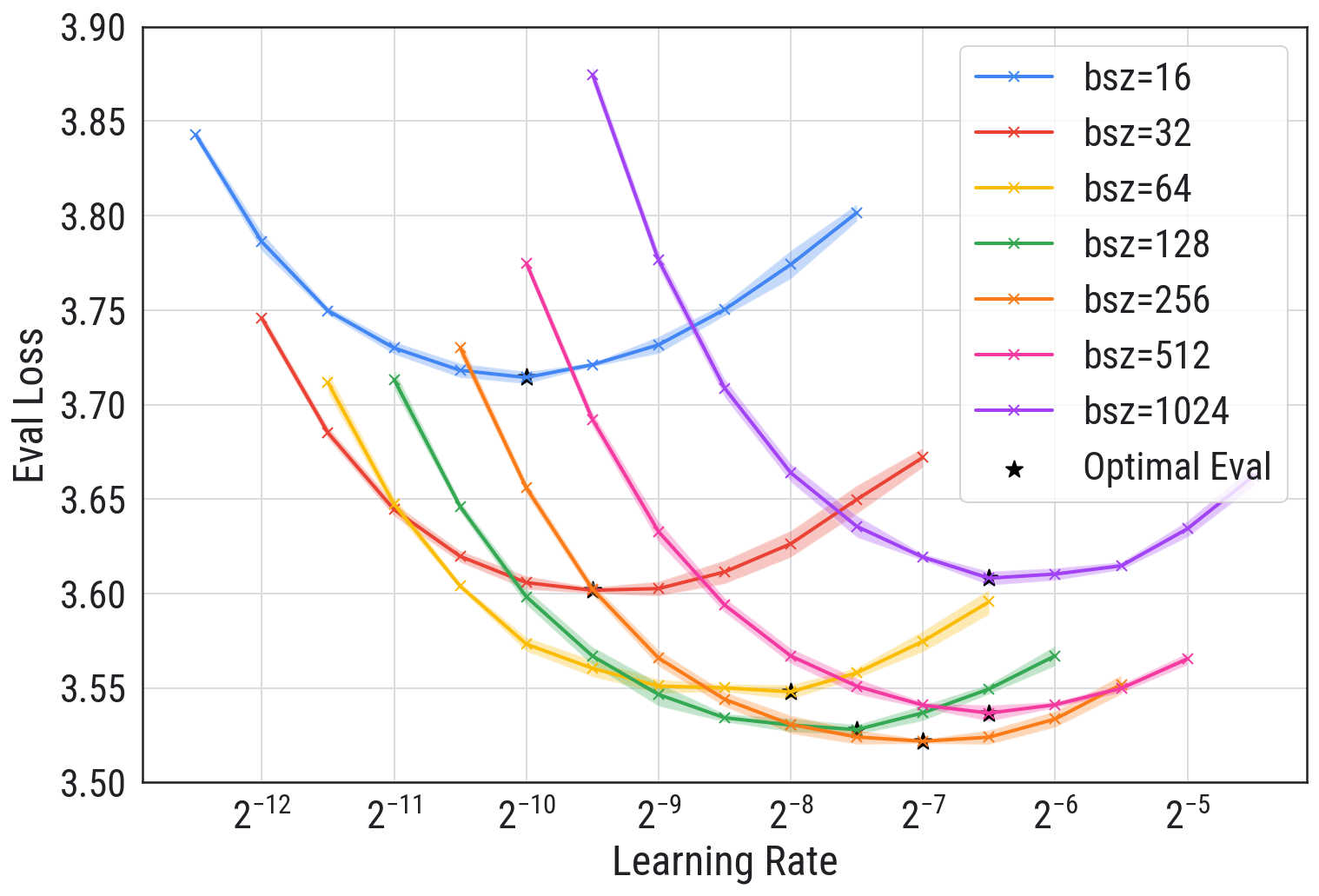}
         \caption{Model Size 19M}
         \label{fig:batch size vs learning rate 19m.}
     \end{subfigure}
     \hfill
     \begin{subfigure}[h]{0.3\textwidth}
         \centering
         \includegraphics[width=\textwidth]{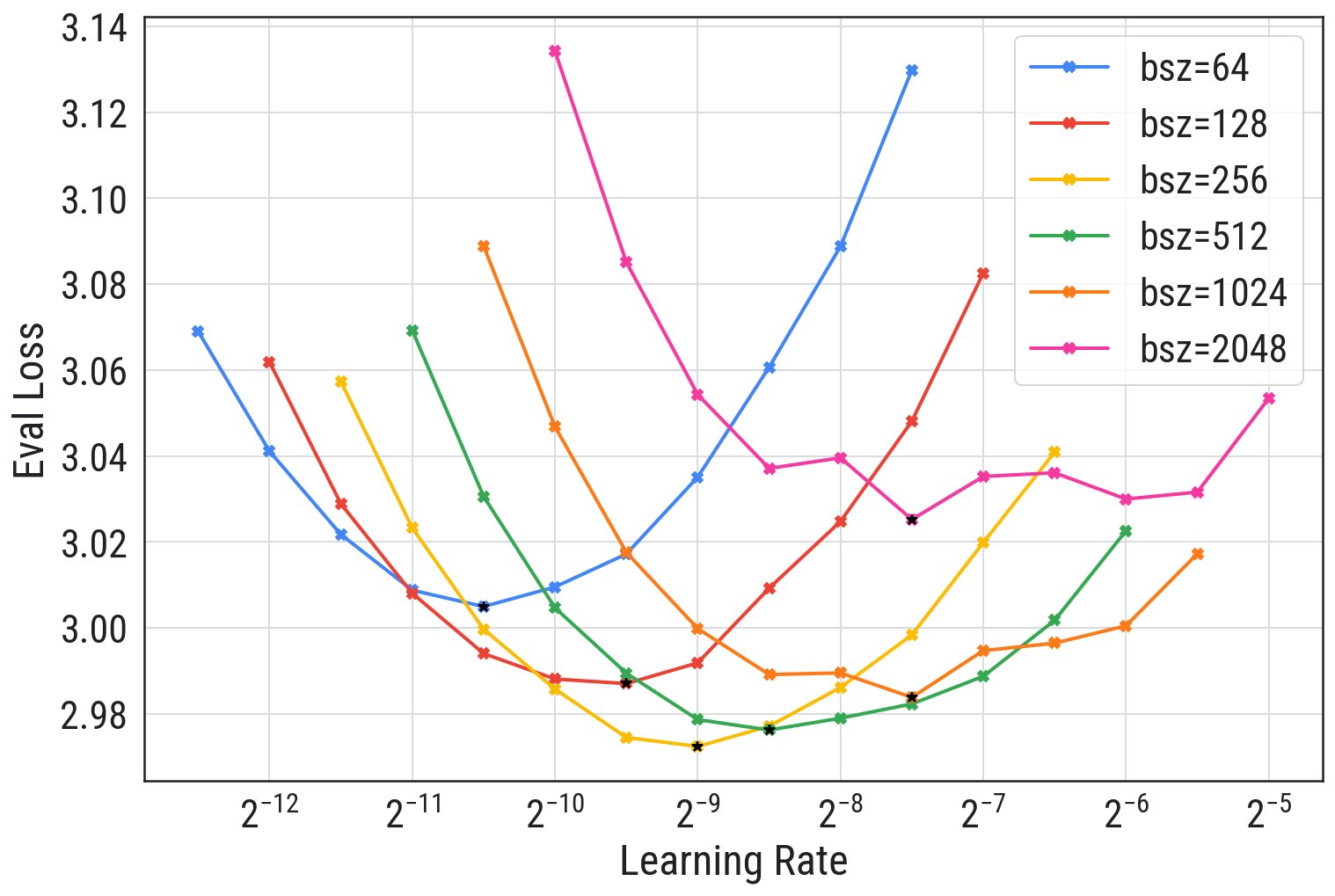}
         \caption{Model Size 151M}
         \label{fig:batch size vs learning rate 151m.}
     \end{subfigure}
     \hfill
     \begin{subfigure}[h]{0.3\textwidth}
         \centering
         \includegraphics[width=\textwidth]{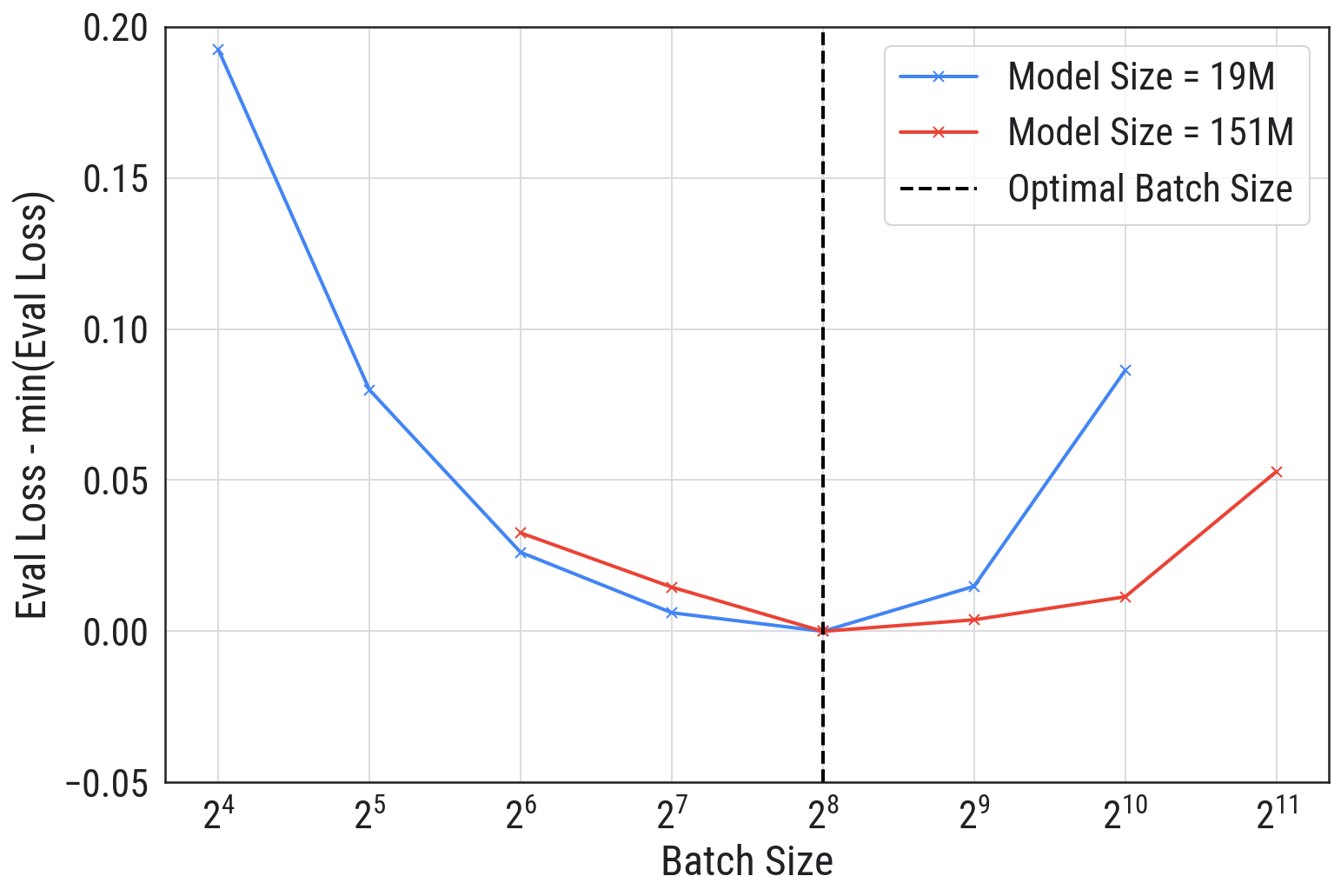}
         \caption{Optimal Batch Size}
         \label{fig:perf vs learning batch size}
     \end{subfigure}
     \hfill
    %  \begin{subfigure}[h]{0.3\textwidth}
    %      \centering
    %      \includegraphics[width=\textwidth]{figures/May9/approach12-chinchilla.png}
    %      \caption{Optimal Model Size}
    %      \label{fig:five over x}
    %  \end{subfigure}
        \caption{{\bf Optimal Batch Size.} Learning rate search for various batch sizes. (a) 19M model,
        (b) 151M model. (c) Performance as a function of batch size is U-shaped and is sensitive to the choice of batch size. 
        Both large and small batch sizes lead to sub-optimal performance.
        }
        \label{fig:batch-size experiments}
\end{figure}

\paragraph{Experimental Results.}  Fig.~\ref{fig:batch size vs learning rate 19m.} and Fig.~\ref{fig:batch size vs learning rate 151m.} illustrate the evaluation loss as a function of learning rate for various batch sizes. Aggregating the best evaluation loss for each batch size, we observe a clear U-shaped curve in Fig.~\ref{fig:perf vs learning batch size}. This demonstrates that both excessively small and large batch sizes can negatively impact model performance.

\paragraph{Discussion.} The conventional wisdom that smaller batch sizes lead to better performance may not always apply to language model pretraining \citep{vyas2023beyond}. While gradient noise from small batch sizes can potentially benefit generalization, it may also impede optimization. The observed U-shaped relationship between loss and batch size raises interesting questions. Why does this U-shape occur, and what trade-offs determine the optimal batch size? These questions remain open for further investigation and could provide valuable insights into improving the efficiency of large-scale language model training.

\subsection{Discussion}
Through three examples, we provided evidence that regularization, either implicit or explicit, may not be necessary for language model pretraining. It may no longer be the main driving principle for understanding pretraining or making informed decisions during training. This raises a crucial question: what are the emerging guiding principles in the scaling-centric paradigm?  See Section \ref{sec:discussion-model-comparison} for a more in-depth discussion.

\section{Scaling Law Crossover, a Curse from Scale?}\label{sec:no-free-lunch}

In the generalization-centric paradigm, the scales we operate on are significantly smaller than those in the LLM setting. This allows us to test new ideas on smaller datasets like CIFAR10 and, if successful, scale them up to ImageNet. Since hyperparameter tuning is feasible even at ImageNet scales, we can readily test our ideas without excessive concern about hyperparameters. Consequently, when proposing new ideas, we often pay little attention to how hyperparameters should evolve with scale, assuming users can perform the tuning themselves. For instance, when introducing alternative architectures (e.g., skip connections, batch normalization), data augmentation techniques, or new optimizers like AdamW, we don't necessarily need to provide guidance on how the learning rate and other hyperparameters should scale with model and dataset sizes.

However, in the ``skydiving'' regime, the sheer scale of data and models presents a significant challenge. Traditional hyperparameter tuning becomes impractical due to the immense computational costs involved. This makes it incredibly difficult and expensive to verify new ideas or compare different approaches at scale. Consider the seemingly simple question of whether a global gradient clipping norm of 1 or 2 is more effective in a 100B parameter model. Answering this question through direct experimentation would require substantial resources and time, highlighting the unique challenges posed by this regime.

In what follows, we present a new phenomenon, termed  {\it scaling-law crossover}, where the effectiveness of different techniques reverses at a certain scale. One idea outperforms another below a critical scale, while the opposite holds above it.

We present three cases of scaling-law crossover with increasing complexity, offering explanations for the first two but leaving the third as an open question that underscores the complexities of scaling. Unlike the traditional ``test on CIFAR, scale to ImageNet'' workflow, the reality of scaling laws may necessitate a continuous climb up the scaling ladder: testing at progressively larger scales until a crossover is observed, or we gain enough confidence to bet that the idea is effective at the desired scale. Therefore, evaluating an idea's potential may take days or even weeks, and require 100+ GPUs and a group of researchers. This process becomes increasingly harder and more costly as we climb up the scaling ladder.

\subsection{Warmup: Training Instability}
\label{sec:warmjup-instability}

\begin{figure}[t!]
     \centering
     \begin{subfigure}[h]{0.3\textwidth}
         \centering
         \includegraphics[width=\textwidth]{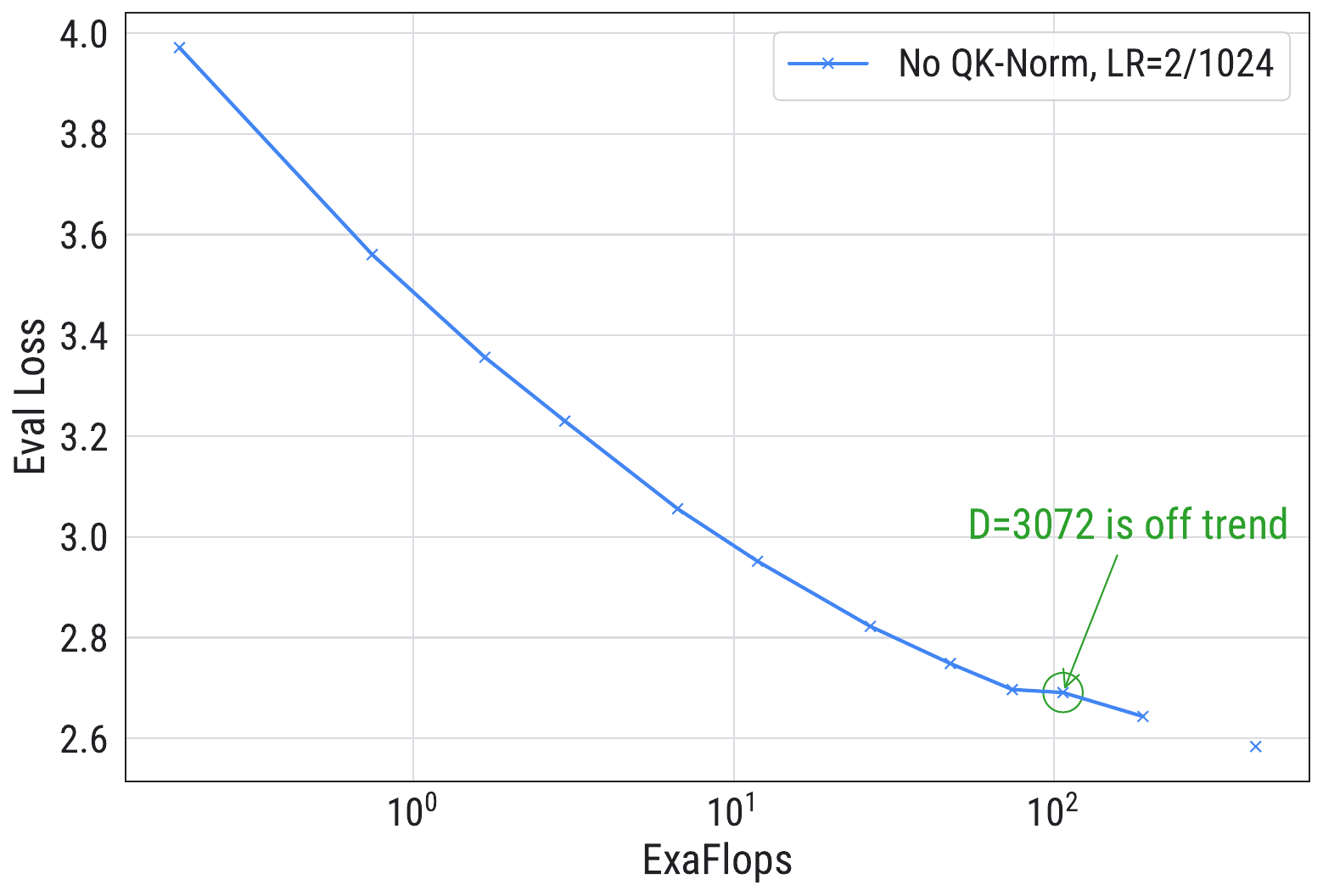}
         \caption{Original Proposal}
         \label{fig:og-proposal.}
     \end{subfigure}
    %  \hfill
     \begin{subfigure}[h]{0.3\textwidth}
         \centering
         \includegraphics[width=\textwidth]{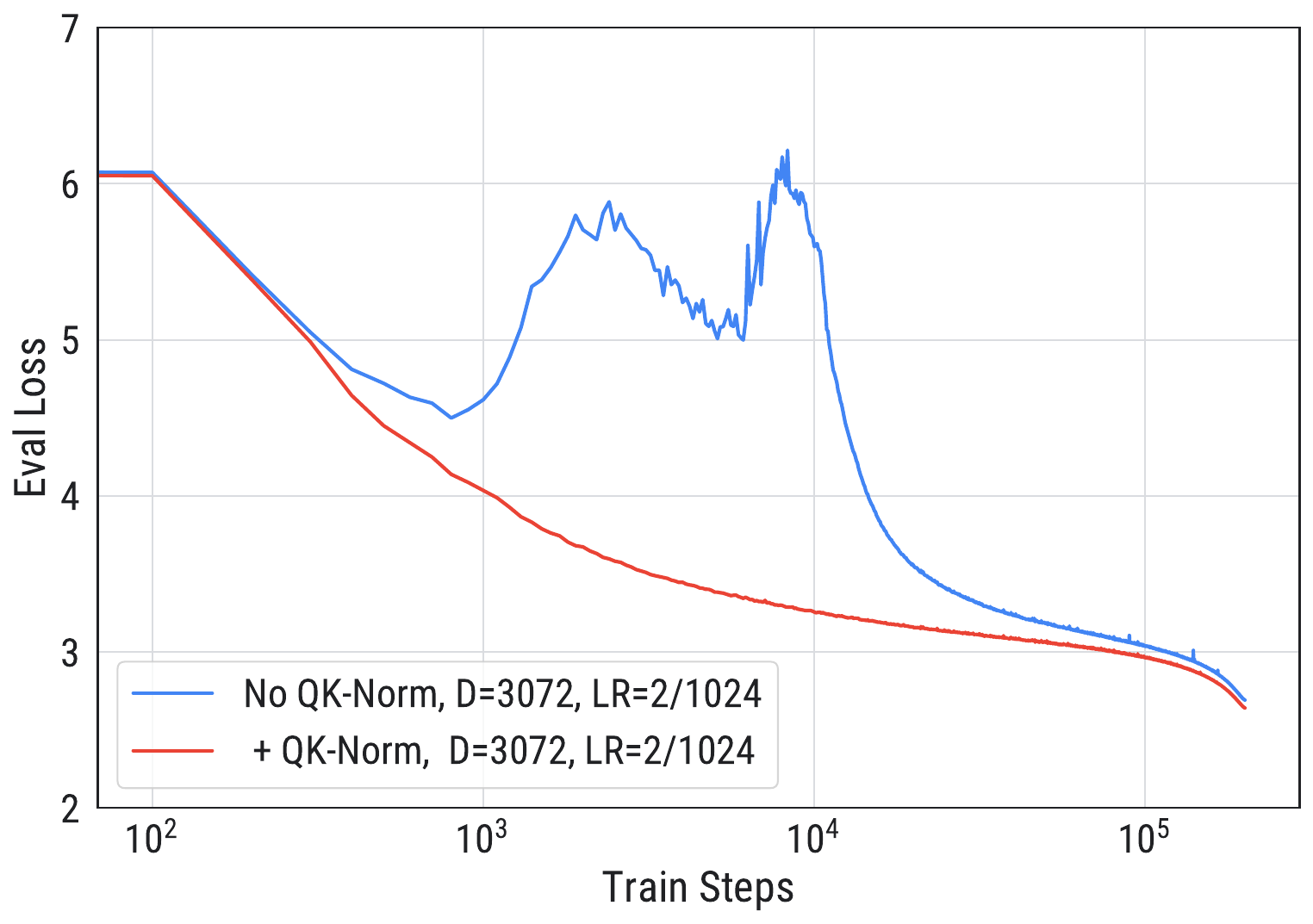}
         \caption{QK-Norm fixes Instability.}
         \label{fig: qk-fix-instability.}
     \end{subfigure}
    %  \\
\begin{subfigure}[h]{0.3\textwidth}
         \centering
     \includegraphics[width=\textwidth]{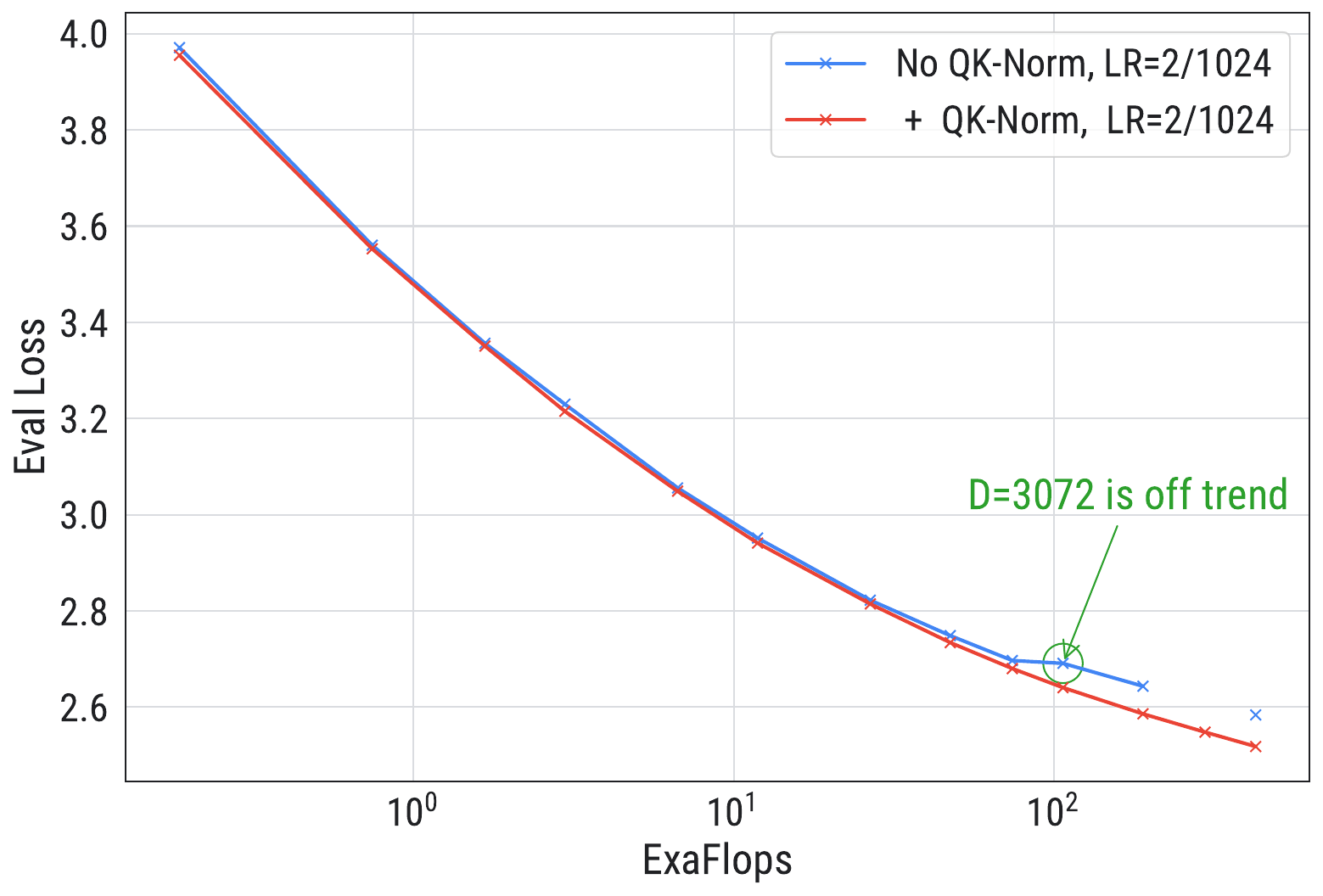}
    %  \hfill
     \caption{QK-Norm Improve Perf}
         \label{fig:qk-improve-perf.}
     \end{subfigure}
        \caption{{\bf Training Instability.} 
    %   Constant learning rates perform well at smaller scales but deteriorate at larger scales. This is evident in two scenarios. 
(a) The model with $D=3072$ deviates from the expected scaling trend.
(b) Closer examination of the training dynamics reveals that the $D=3072$ model exhibits instability. This instability is mitigated by the application of QK-Norm.
(c) With QK-Norm, the scaling law exhibits a normal and smooth behavior.
}
        \label{fig:crossover-learing-rate}
\end{figure}

Training instability in large language models (LLMs) is a widely recognized challenge within the research community \citep{liu2020understanding, chowdhery2023palm, pmlr-v202-dehghani23a, zhang2023opt,molybog2023theory, cohen2022adaptive, wortsman2023small}. In this section, we reproduce this phenomenon and explore potential solutions. While our initial fix appears to resolve the issue, it hides a deeper problem in scaling strategies.

\paragraph{Experiment Setup.} We train a series of models with varying dimensions $(D, F) = (128k, 512k)$ for $k \in [1, 2, 3, 4, 6, 8, 12, 16, 20, 24, 32, 40, 48]$. All models share the same batch size (256), number of layers (6), sequence length (512), and training steps (200,000, not Chinchilla-Optimal). We use a constant learning rate $\text{LR}=2/1024$ for all model sizes.  

In Figure \ref{fig:og-proposal.}, we plot the evaluation loss against computational cost (flops) for each model. While performance generally improves with scale, a discontinuity emerges around $10^2$ exaflops ($D=3072$). Examining the learning dynamics for $D=3072$ (Figure \ref{fig: qk-fix-instability.}) reveals training instability in larger models: the evaluation loss spikes during training (between 1,000-10,000 steps) but self-corrects later. Clearly, our scaling approach requires adjustment.

Fortunately, several techniques exist to address training instability: Z-loss \citep{chowdhery2023palm}, extended warmup periods \citep{wortsman2023small}, learning rate reduction, increased weight decay, and QK-Norm (to prevent attention logit explosion \citep{pmlr-v202-dehghani23a}).

We try QK-Norm, and it proves effective. Not only does it resolve the instability for $D\geq 3072$ (Figure \ref{fig: qk-fix-instability.}), but it also enhances performance for larger model sizes (Figure \ref{fig:qk-improve-perf.}). However, a crucial question remains: have we truly addressed the underlying issue, or have we merely patched a symptom?

\subsection{Sub-Optimal Learning Rate Scaling Rule}
\label{sec:sub-optimal learning rate}

Years of research investment in understanding the training dynamics of neural networks has taught us a valuable lesson: learning rates should be adjusted based on training specifics, particularly model scale \cite{goh2017why, lee2019wide, sohl2020infinite, xiao2019disentangling, yang2022tensor, bi2024deepseek}. The challenge lies in determining how to adjust them effectively.

We compare two proposals: using a constant learning rate (LR = 2/1024 as in Section \ref{sec:warmjup-instability}) versus scaling the learning rate with model width ($\text{LR} = 2/D$) \cite{everett2024scaling}.

\underline{Proposal Blue}. Constant learning rate $\text{LR}=2/1024$ for all model sizes.  

\underline{Proposal Red}. Learning rate scales with model dimension,  $\text{LR}=2/D$. 

We evaluate these proposals with and without QK-Norm. Results are presented in Figure \ref{fig:crossover-learing-rate}.

\begin{figure}[t!]
     \centering
     \begin{subfigure}[h]{0.35\textwidth}
         \centering
         \includegraphics[width=\textwidth]{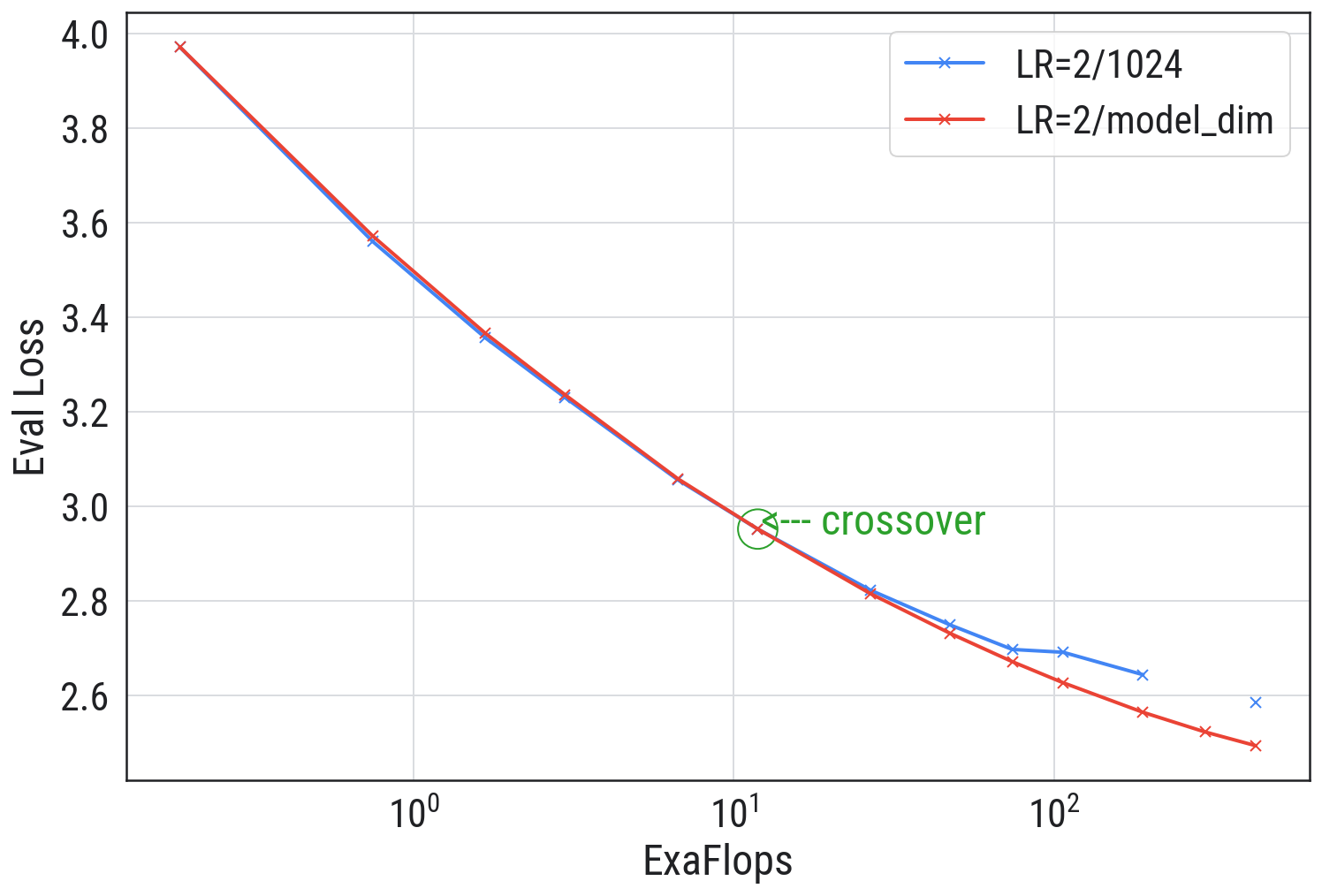}
         \caption{No QK-Norm}
         \label{fig:no-qk-crossover.}
     \end{subfigure}
\begin{subfigure}[h]{0.35\textwidth}
         \centering
     \includegraphics[width=\textwidth]{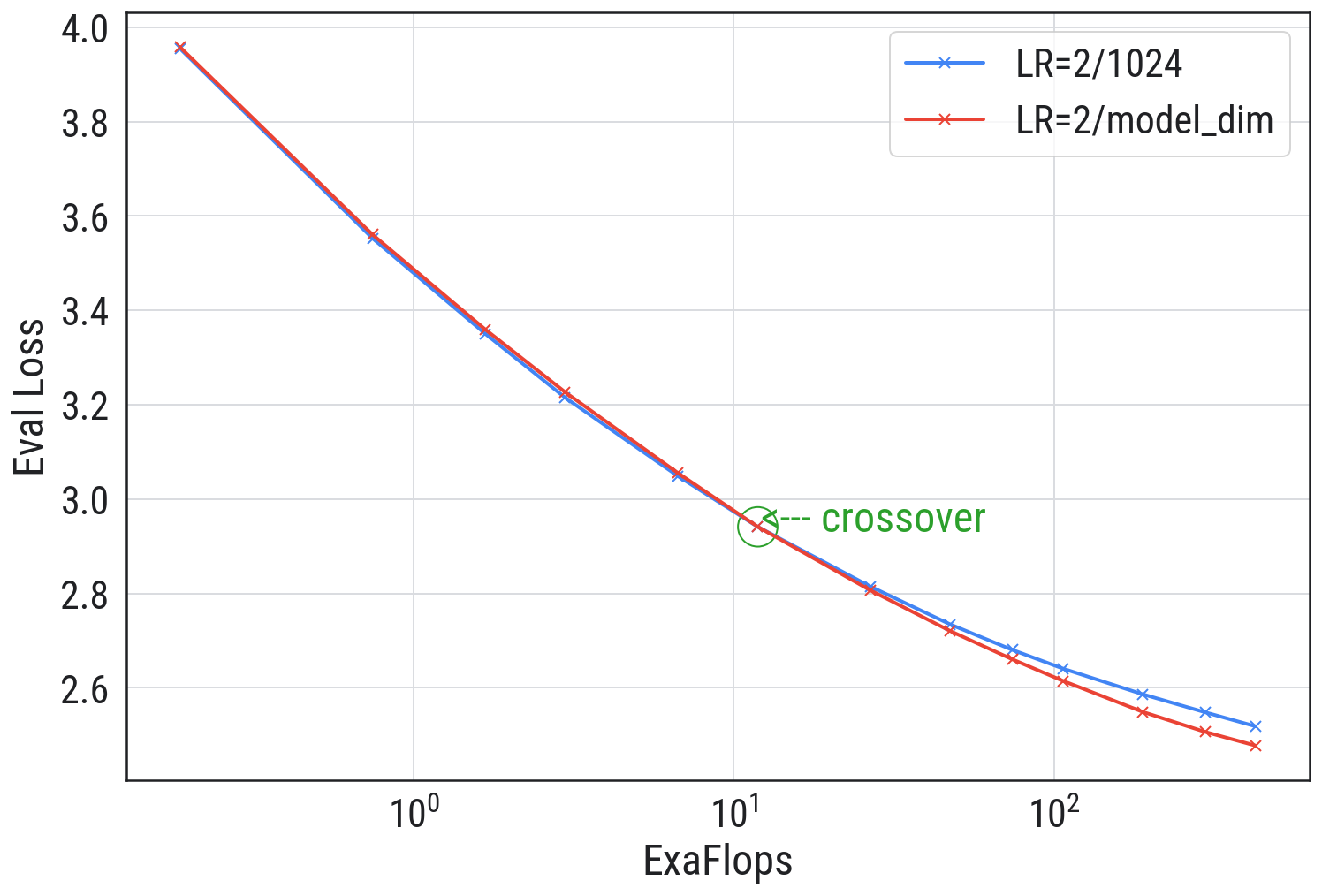}
    %  \hfill
     \caption{With QK-Norm}
         \label{fig:with qk crossover.}
     \end{subfigure}
        \caption{{\bf Constant Learning Rate vs $\text{LR}=2/D$.} 
    %   Constant learning rates perform well at smaller scales but deteriorate at larger scales. This is evident in two scenarios. 
A constant learning rate $\text{LR} = 2/1024$ and a learning rate of $\text{LR} = 2/D$ yield nearly identical performance at small scales ($D \leq 1024$). However, their performance diverges at larger scales ($D > 1024$), with $\text{LR} = 2/D$ demonstrating superior performance both (b) with and (a) without QK-Norm.
}
        \label{fig:crossover-learing-rate-new}
\end{figure}

\paragraph{Observations.}  
\begin{enumerate}
    \item Using $2/D$ learning rates mitigates training instability, even without QK-Norm. 
    \item 
    While the two proposals yield nearly identical performance at small scales ($D \leq 1024$), with \underline{Proposal Blue} potentially slightly better, a crossover occurs at $D \geq 1024$. Beyond this point,  \underline{Proposal Red} consistently outperforms  \underline{Proposal blue}, and this performance gap widens with increasing scale, regardless of the presence or absence of QK-norm (Fig.~\ref{fig:no-qk-crossover.} and \ref{fig:with qk crossover.}.)
\end{enumerate}

Retrospectively, the compute inefficiency of \underline{Proposal Blue} can be attributed to sub-optimal learning rate or sub-optimal hyperparameter choices. However, these inefficiencies may not be apparent until we identify the sub-optimal component in our proposal and discover a new one to fix it (in this case, changing learning rate scaling rule from constant to $2/D$). We might have mistakenly believed that adding QK-norm to \underline{Blue} fully addresses the issue, unaware of deeper problems within our proposals. This raises three questions:
\begin{itemize}
\item Is the $2/D$ learning rate scaling rule a true cure or merely a band-aid solution? Could there be an even better one?
\item Are there inherent limitations or hidden bottlenecks in our model architectures that degrade performance at large scales but are not observable at small scales?
\item How can we effectively distinguish between proposals whose performance is nearly indistinguishable at small scales, yet exhibits significant differences at larger scales that are too expensive to run?
\end{itemize}

\subsection{Sub-optimal Weight Decay Scaling Rule}
\label{sec: weight-decay suboptimal}

\paragraph{Experiment Setup.}  The architectures remain consistent with previous sections, varying only the scaling factor $k$ in $(D, F) = (128k, 512k)$. Crucially, models are trained to Chinchilla-Optimal, meaning the training horizon scales with model size (20 tokens per parameter).

\underline{Proposal Blue}:  We employ muP (Maximal Update Parametrization \citep{yang2022tensor}) and a constant, dimension-independent weight decay. Optimal hyperparameters were determined through hyperparameter search with independent weight decay ($\lambda=0.000566)$ which remains the same for all $D$, and learning rate ($\eta=0.0055$) for the base model $D=512$.

\underline{Proposal Red}:  We scale the learning rate inversely with the model size ($\text{LR}=2/D$), and additionally, the weight decay is scaled similarly, with independent weight decay $\lambda =0.1 \times 2/D$.

% We scale the learning rate with $2/D$ and, in addition, the independent weight decay is scaled like $0.2 /D$. 

Note that \underline{Proposal Blue} is consistent with the recommendation from the muP paper ``weight decay should scale independently with width'' \citep{yang2022tensor, yang2023tensor}. 

\begin{figure}[t!]
     \centering
     \begin{subfigure}[h]{0.3\textwidth}
         \centering
         \includegraphics[width=\textwidth]{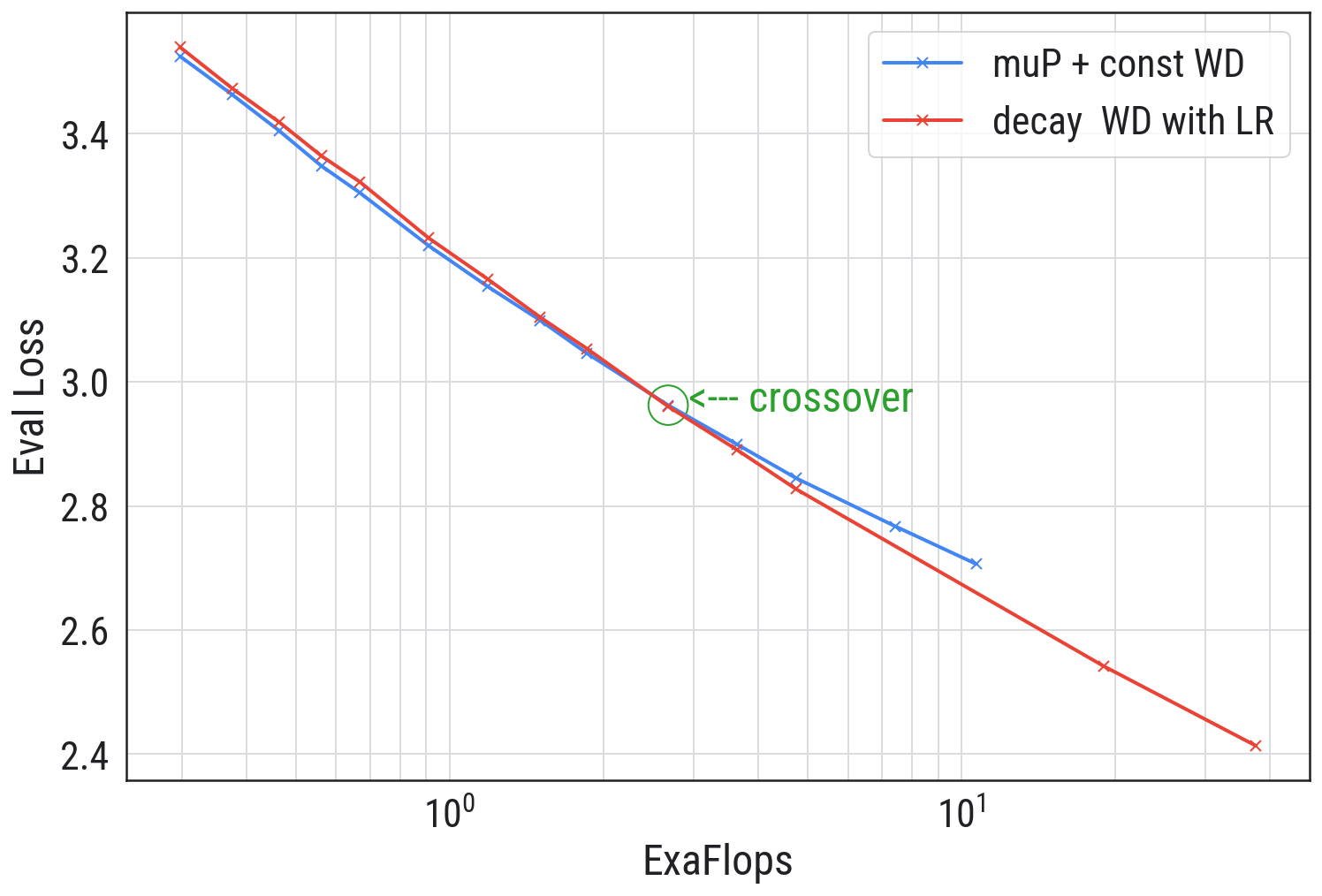}
         \caption{Vary Weight Decay Scaling}
         \label{fig:weight decay scaling.}
     \end{subfigure}
     \hfill
     \begin{subfigure}[h]{0.3\textwidth}
         \centering
         \includegraphics[width=\textwidth]{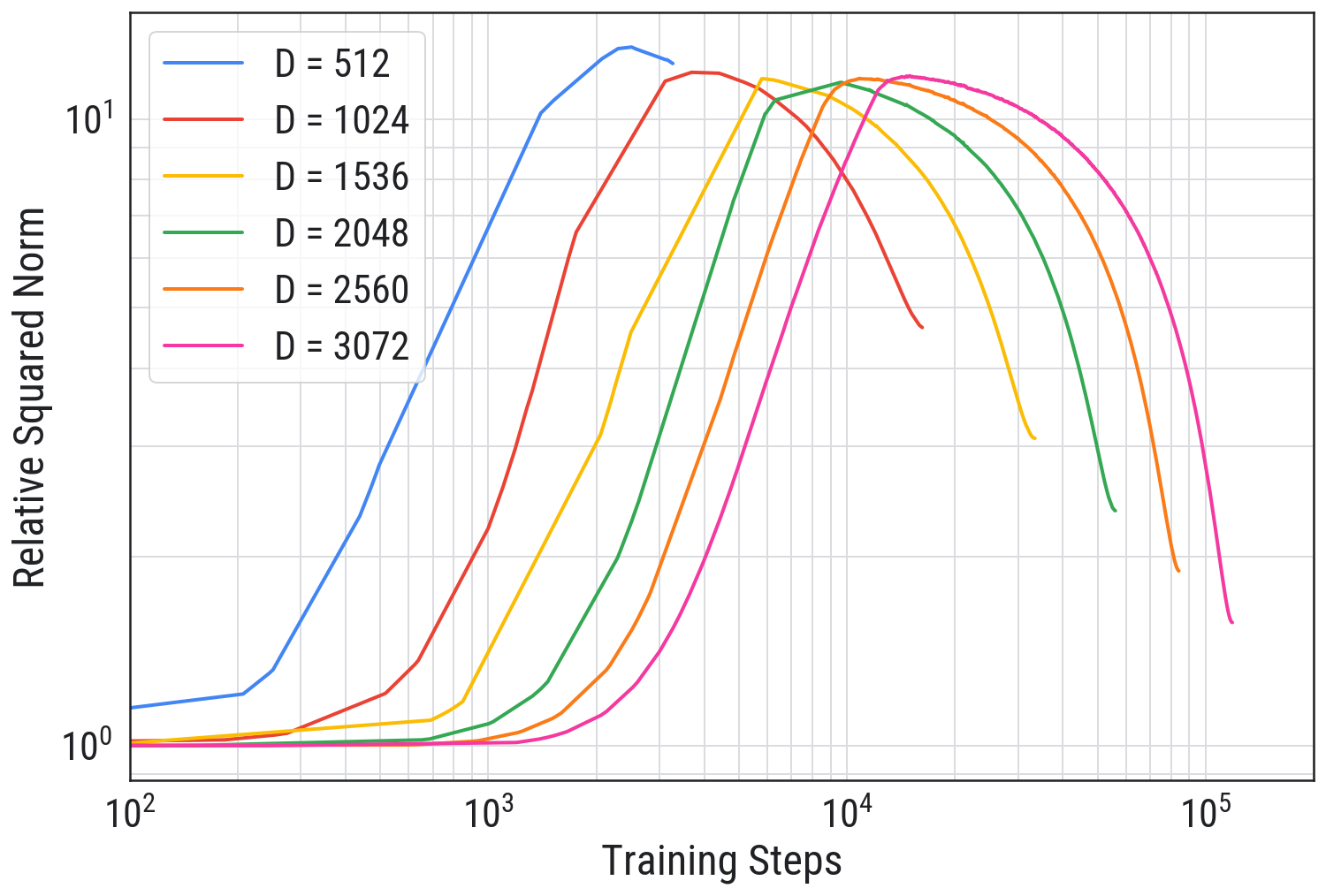}
         \caption{Constant Weight Decay}
         \label{fig:const weight decay.}
     \end{subfigure}
     \hfill
     \begin{subfigure}[h]{0.3\textwidth}
         \centering
         \includegraphics[width=\textwidth]{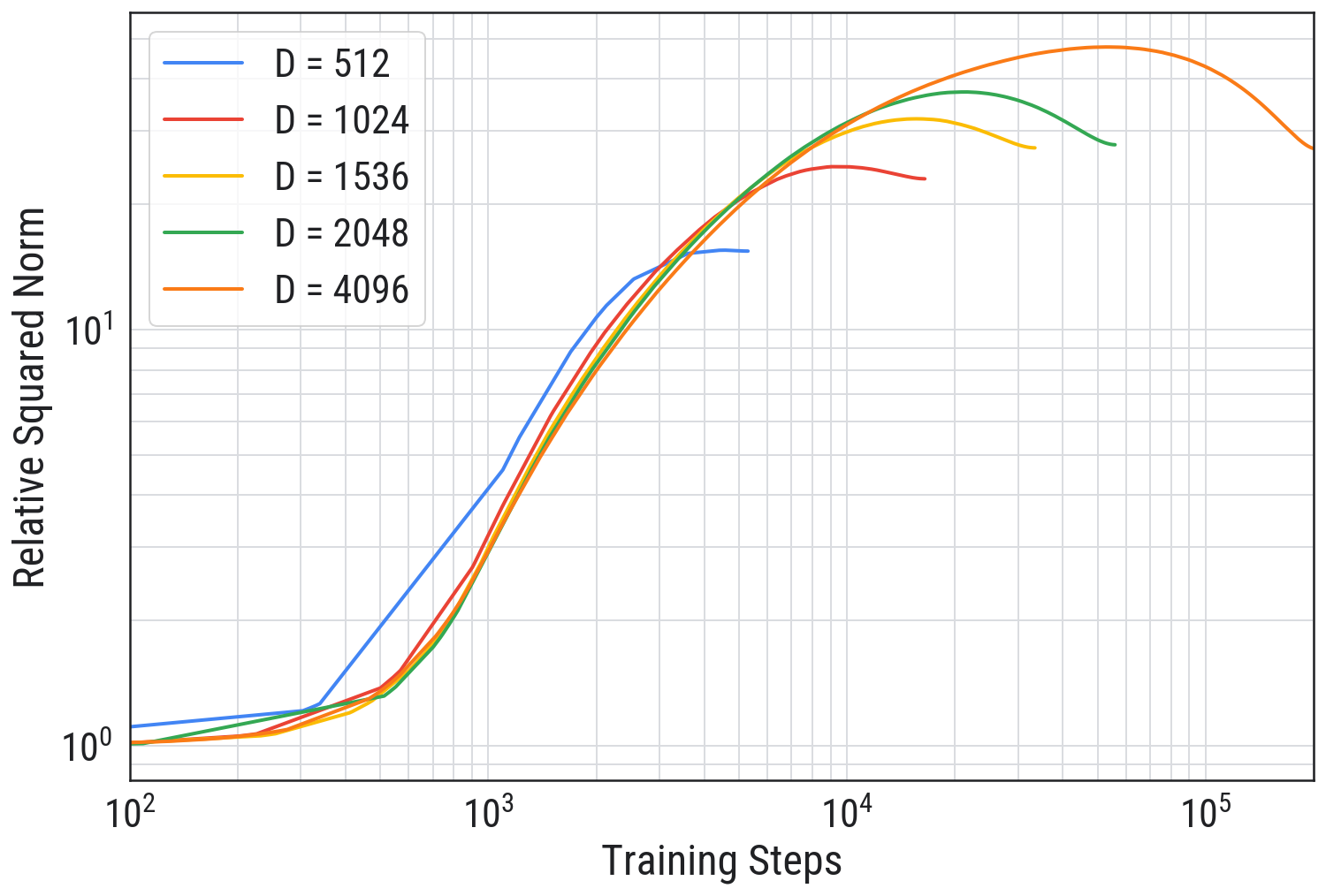}
         \caption{Decaying Weight Decay}
         \label{fig:decaying weight decay.}
     \end{subfigure}
     \hfill
       \caption{\textbf{Crossover Phenomenon due to Sub-optimal Weight Decay.} (a) A constant weight decay strategy initially outperforms a decaying weight decay approach at smaller scales. However, this advantage diminishes and reverses at larger scales, demonstrating a crossover phenomenon. (b) With constant weight decay, parameter norms are significantly suppressed throughout training. (c) Decaying the weight decay alongside the learning rate results in less suppressed parameter norms.}
        \label{fig:weight decay experiments}
\end{figure}

\paragraph{Observation.}  While muP is often considered a best practice for scaling up models, its performance gains observed at smaller scales may not translate effectively to larger ones. A crossover point emerges, beyond which \underline{Proposal Blue} (based on muP) loses its advantage; see Fig.~\ref{fig:weight decay scaling.}. This is likely because muP assumes that the number of training tokens (and thus, training steps) is constant with respect to model dimension ($D$). However, this assumption doesn't hold in practice, especially with Chinchilla-Optimal scaling, which suggests a training token count of $20\mathscr{N}$ (where $\mathscr{N}$ is the number of parameters, proportional to $D^2$ for fixed layers). Therefore, muP's underlying assumption conflicts with the Chinchilla-Optimal scaling strategy \citep{everett2024scaling}. 
As shown in Fig.~\ref{fig:const weight decay.}, constant weight decay leads to significant suppression of parameter norms throughout training. Scaling the weight decay with the learning rate ($\text{LR} = 2/D$) results in better parameter norm dynamics.

\subsection{Is Gradient Normalization a Good Idea?}
\label{subsection:is-normaliation-a-good-idea}
% Let's explore new ideas to improve scaling law. 

\paragraph{Experiment setup.} In this experiment, we co-scale model dimension $D$ and number of layers $L$ with a fixed aspect ratio $D/L = 512/6$ and train all models to Chinchilla-Optimal. Our learning rate search procedure suggests the formula $\eta = 2 / D\times (6/N)^{0.675} \times 2^{0.25}$. We call this \underline{Proposal Red}.

\begin{figure}[b!]
     \centering
     \begin{subfigure}[h]{0.4\textwidth}
         \centering
         \includegraphics[width=\textwidth]{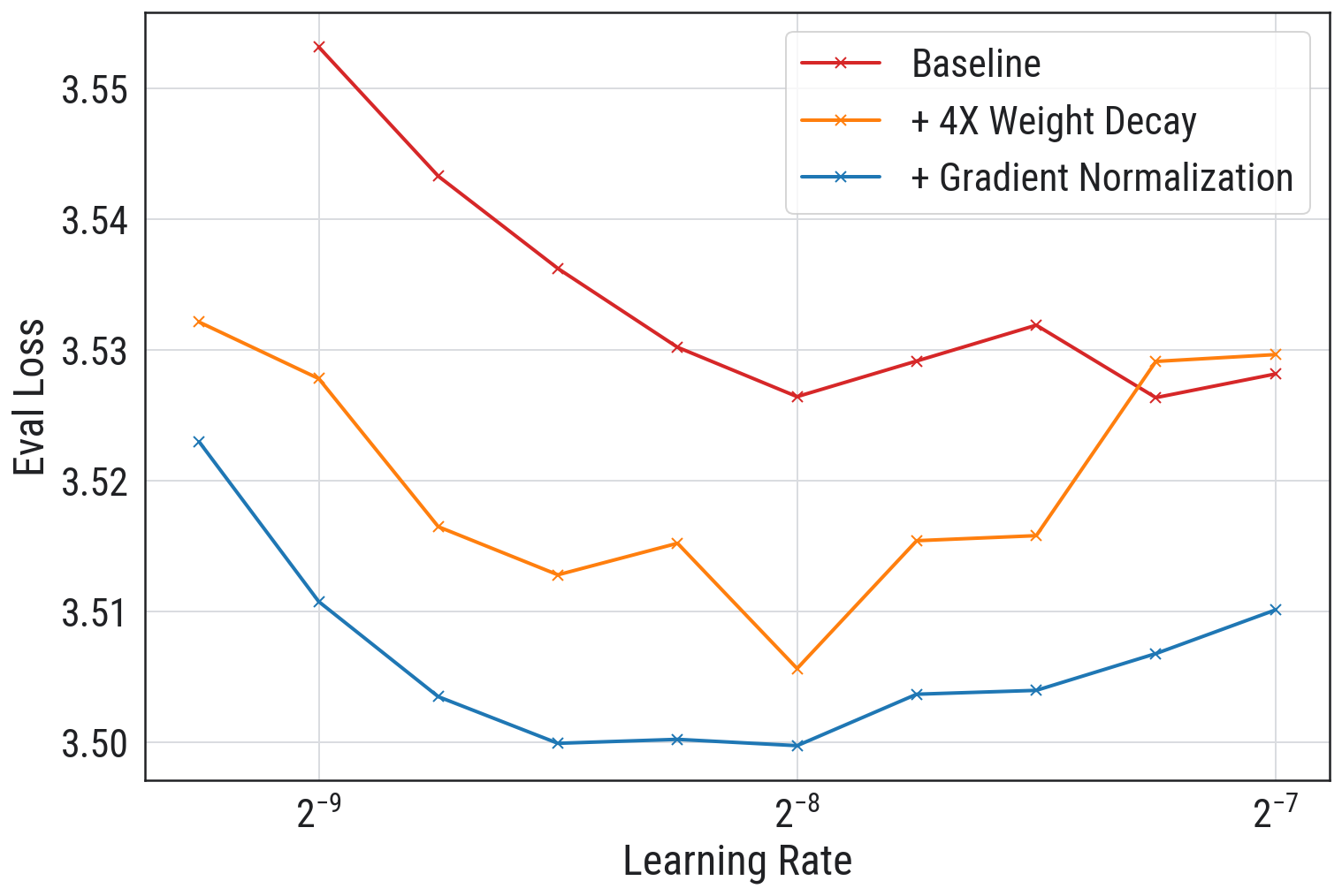}
         \caption{Ablation Study.}
         \label{fig:ablation of gradient and weight decay}
     \end{subfigure}
    %  \hfill
     \begin{subfigure}[h]{0.4\textwidth}
         \centering
         \includegraphics[width=\textwidth]{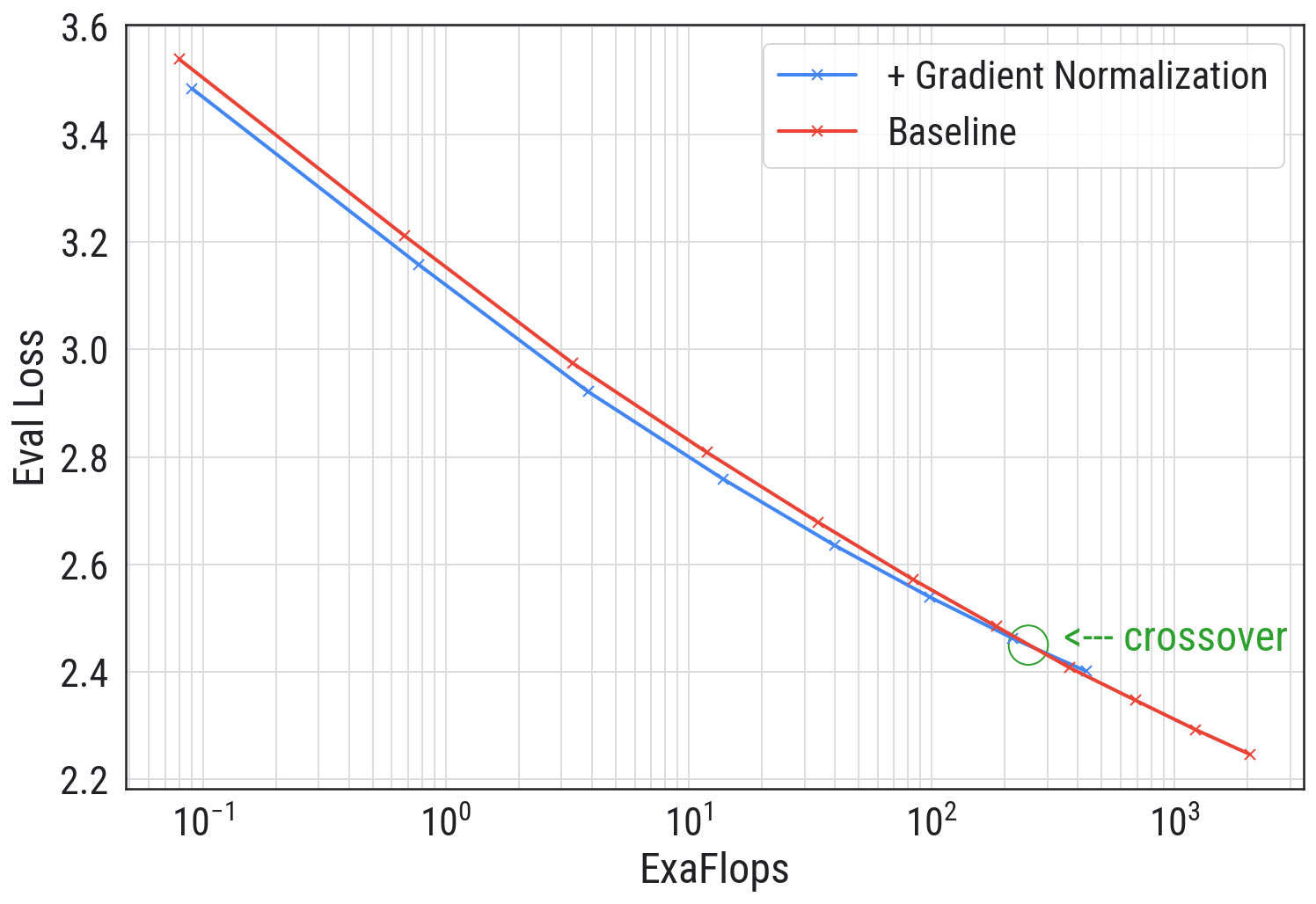}
         \caption{Gradient Normalization and Crossover}
         \label{fig:crossover-gradient-normalization.}
     \end{subfigure}
     \hfill
        \caption{{\bf Gradient Normalization Leads to Scaling Law Crossover?} (a) An ablation study showing \underline{Proposal Blue} performs better at small scales. In particular, gradient normalization not only improves performance but also reduces learning rate sensitivity in the scale we are testing. 
        (b) Performance gain from \underline{Proposal Blue} does not transfer to large scale. A crossover occurs around $2-3 \times 10^2$ exaflops. 
        }
        \label{fig:gradient normalization fig}
\end{figure}

\underline{Proposal Blue:} We propose the following modifications to the base model:

\begin{enumerate}
\item \textbf{Activation Function and MLP Dimension:} Replace the GeLU activation function with its gated variant, GeGLU \citep{shazeer2020glu}, and increase the hidden dimension of the MLP to $F = 6D$. Gated activations like GeGLU are commonly believed to improve model performance.

\item \textbf{Weight Decay:} Increase the weight decay by a factor of 4. 

\item \textbf{Gradient Normalization:} Normalize the raw gradients by their root mean square (RMS) before passing them to the AdamW optimizer.
\end{enumerate}

The ablation study (Fig.~\ref{fig:ablation of gradient and weight decay}) shows promising results for these modifications, particularly in terms of improved performance and reduced learning rate sensitivity \citep{wortsman2023small}. 
It seems we may have found an innovation. 
Let's now evaluate \underline{Proposal Blue} at scales.

\paragraph{Observation.}  
Evidently, we encounter another scaling law crossover, as illustrated in Figure \ref{fig:crossover-gradient-normalization.}. While \underline{Proposal Blue} initially exhibits promising results at smaller scales, this advantage diminishes, leading to a crossover point around $2-3 \times 10^2$ exaflops --- a significantly larger scale than observed in previous experiments. Unlike those instances, the cause of this crossover remains elusive. Does gradient normalization become inherently detrimental at larger scales? Should the learning rate scaling formula be adjusted when incorporating gradient normalization?  Even if we successfully address the limitations of the gradient normalization proposal and verify its effectiveness up to $10^3$ exaflops, can we confidently assert that the fix will remain effective at even larger scales, such as $10^4$ exaflops?

\subsection{Discussion} 
We discuss several direct consequence of scaling law crossover and leave a more in-depth discussion for the next section. 

\begin{enumerate}
    \item {\bf Hyperparameter Tuning at Scale.} As performance is highly sensitive to the choice of hyperparameter scaling rules, (False Negative) good ideas may be killed owing to insufficient hyperparameter tuning and (False Positive) sub-optimal ideas may be promoted due to weak baseline. 
    % We therefore recommend that researchers report not only their ideas but also their specific scaling strategies and hyperparameter choices (i.e., their ``scaling recipes''); see e.g., \cite{yang2022tensor, everett2024scaling}.
    \item {\bf Credit Assignment.} 
    The crossover scaling phenomenon underscores that demonstrating impressive performance at small scales is insufficient. While proposing new ideas and testing them at small scales remains crucial, rigorously verifying ideas at large scales demands substantial effort, resources, and, crucially, faith in their potential. Thus, we argue that scaling up existing ideas and rigorously demonstrating their effectiveness at scale is as important as, or even more important than, proposing new ideas and testing them on small scales. Both types of contributions are essential and should be recognized and valued.

    \item {\bf Avoid Biased Search Spaces.}
    The scaling law crossover phenomenon indicates that 
    it is crucial to avoid overemphasizing ideas that work well at small scales. This narrow focus might lead us to miss groundbreaking approaches, similar to the ``vision transformer'' \citep{dosovitskiy2020image}, that excel at large scales but might not shine on smaller ones.
    
\end{enumerate}

\section{Discussion and Conclusion}\label{sec:discussion-model-comparison}
Let us revisit the two central themes: model comparison at scale and guiding principles for scaling.

\subsection{Model Comparison at Scale}

The ability to effectively and reliably compare models is fundamental to advancing machine learning.
In the generalization-centric paradigm, the validation set approach remains a {\it simple}, {\it reliable}, {\it cost-effective} and {\it theoretically grounded} method for such comparisons. However, this approach does not apply to the scaling-centric paradigm. The immense scale often prevents training multiple models for comparison. Furthermore, the phenomenon of scaling law crossover — where the relative performance of methods can change as models scale — poses a fundamental challenge: we cannot simply compare two models at small scales and assume that the observed ranking will hold at larger scales. It raises a fundamental question: 

\textbf{Model Comparison at Scale}: How to compare models at a scale where training is feasible only once?

We discuss two possible methods below.

\subsubsection{Scaling Law Extrapolation for Model Comparision}
\begin{figure}[t!]
    \centering
    \includegraphics[width=0.48\textwidth]{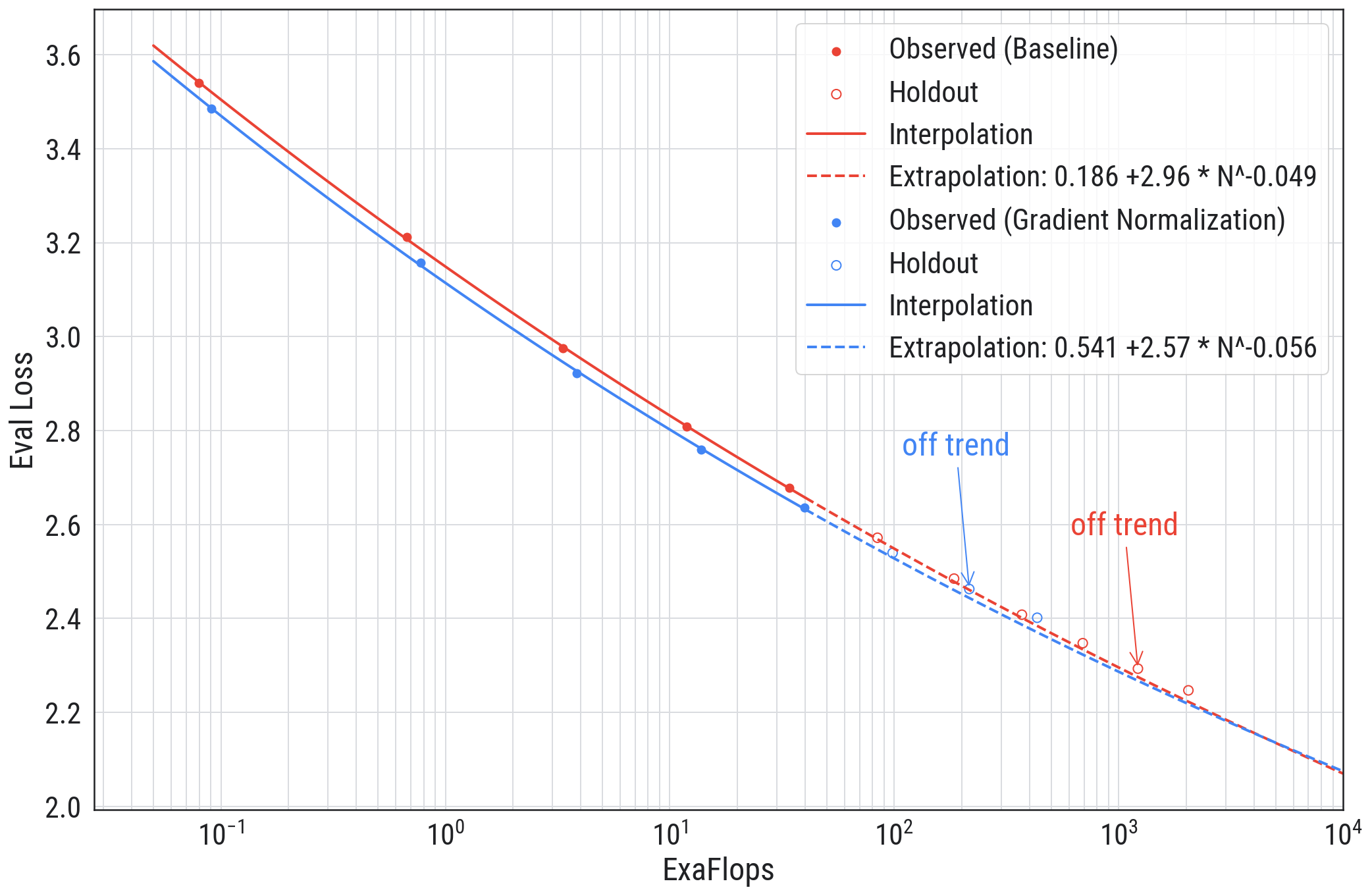}
    \caption{{\bf Scaling Law Extrapolation}:  Naively fitting a power law to observed data points can lead to inaccurate extrapolations.  For example, our ``Blue'' and ``Red'' proposals showed significant deviations after 10x and 30x compute extrapolation, respectively. 
    }
    \label{fig:scaling-law-prediction}
\end{figure}

% \paragraph{Scaling Law Extrapolation.} 
The first approach relies on scaling law extrapolation \citep{kaplan2020scaling}: extrapolating observations from smaller scales to predict performance at larger scales. Specifically, we assume the following functional form relating loss $\mathscr{L}$ to computational cost (flops, denoted by $f$):
\begin{equation}
\label{eq:scaling-law-prediction}
  \mathscr{L}(f) = a f^{b} + c
\end{equation}
for a given class of models \citep{kaplan2020scaling}, where the parameters $(a, b, c)$ depend on the model's specific characteristics. We then generate a sequence of measurements $\{(f_i, l_i)\}_{1 \leq i \leq k}$ by training a series of models up to a certain scale (e.g., $k=5$ and up to 40 exaflops, as shown in Fig.~\ref{fig:scaling-law-prediction}), where $(f_i, l_i)$ represents the flops and loss of the $i$-th model. Next, we find the optimal values of $(a, b, c)$ that best fit these measurements. Finally, we use Equation \ref{eq:scaling-law-prediction} to make predictions about future performance, such as the expected loss at $f = 5e3$ exaflops.

We apply this approach naively to \underline{Proposal Blue} and \underline{Proposal Red} from Section \ref{subsection:is-normaliation-a-good-idea}, using $k=5$ and $f \leq 40$ exaflops. The results are presented in Figure \ref{fig:scaling-law-prediction}. 

Although our scaling law extrapolations capture the overall scaling trend, the precision of the predictions is insufficient for reliable model comparison.  \underline{Proposal Blue} deviates from the predicted trend around $3 \times 10^2$ exaflops, failing to achieve a 10x extrapolation, while \underline{Proposal Red} deviates around $10^3$ exaflops, failing a 30x extrapolation. Furthermore, the scaling law extrapolations incorrectly predict that crossover occurs in the interval $[4\times 10^3 , 5 \times 10^3]$, while the actual crossover occurs in $[2\times 10^2, 3 \times 10^2]$, as shown in Figure \ref{fig:crossover-gradient-normalization.}. This demonstrates that naively extrapolating scaling laws for model comparison can be unreliable and lacks a solid theoretical foundation. In contrast, the GPT-4 technical report \citep{openai2023gpt} showcases the potential for accurate 1,000x and even 10,000x extrapolations, although the specific techniques and conditions enabling such accurate extrapolations remain unclear.

Overall, to reliably apply scaling law extrapolation for model comparison at scale, we believe extensive research is necessary to fully comprehend both the macro and micro dynamics at play. This may include a deeper understanding of the intricate relationship between optimization, architectures, data, and scales \citep{kaplan2020scaling, bahri2024explaining, hoffmann2022training, deepseekai2024deepseekv2strongeconomicalefficient, paquette2024, bordelon2024dynamical, lin2024scaling}, as well as the subtleties within our machine learning systems. These subtleties encompass factors such as learning rate schedules \citep{hoffmann2022training}, parameter counting choices, the number of warmup steps \citep{porian2024resolving}, curve fitting approaches \citep{besiroglu2024chinchilla}, and even the epsilon value in AdamW \citep{wortsman2023small, everett2024scaling}.

\subsubsection{Are Hyperparameter Transfer sufficient for Model Comparison?}

$\mu$-Transfer~\citep{yang2022tensor} is an important technique for tuning hyperparameters in large models. Its core idea involves parameterizing the network (using maximal update parameterization) and scaling the learning rate appropriately, enabling the zero-shot transfer of optimal hyperparameters (e.g., learning rates, scale of initialization) from small models to much larger ones. While these methods have demonstrated promising results \citep{yang2022tensor, lingle2024large, blake2024u, everett2024scaling}, certain limitations hinder their direct application to model comparison.

Firstly, hyperparameter comparison represents only a small fraction of model comparison, which encompasses optimization choices (e.g., AdamW vs. Lion, schedule-free optimizer, Shampoo \citep{chen2024symbolic, defazio2024road, gupta2018shampoo}), architectural variations (e.g., multi-head attention vs. grouped-query attention \citep{ainslie2023gqa, shazeer2019fast}, transformers vs. non-transformers \citep{gu2023mamba, botev2024recurrentgemma, beck2024xlstm}), and data considerations (e.g., different data mixtures \citep{penedo2024finewebdatasetsdecantingweb}, varying token-to-parameter ratios).

Secondly, the current $\mu$P framework primarily focuses on scaling width\footnote{While there has been some progress in depth scaling~\citep{yang2023tensor, bordelon2023depthwise}, the signal for hyperparameter transfer remains convoluted.} while maintaining other factors like batch size, the number of layers, and training steps as static hyperparameters. This deviates from practical scenarios where batch size, depth, width, and training steps often co-evolve with scale. Empirical evidence suggests that hyperparameters may not transfer seamlessly when scaling more than one dimensions concurrently~\citep{everett2024scaling}.

Thirdly, we may seek quantitative model comparisons rather than just qualitative assessments, where hyperparameter transfer may not be directly applicable. For instance, to reduce inference costs, we might overtrain a small model significantly beyond the Chinchilla-Optimal point and employ grouped-query attention instead of multi-head attention. In this scenario, we would like to quantify and then optimize the computational trade-offs between the overtrained grouped-query model and a Chinchilla-Optimally trained multi-head model. 

In conclusion, hyperparameter transfer technique alone at its current form is not sufficient to resolve the challenge of model comparison at scale. 

\subsection{Guiding Principles for Scaling}

In the pursuit of models that generalize well, regularization plays a central role. It serves as a guiding principle for understanding machine learning algorithms, making informed decisions during training, and inspiring novel ideas. Through the lens of regularization, we can understand a variety of phenomena and discover new techniques in machine learning. We list several examples below:

\begin{itemize}
    \item \textbf{Hyperparameter Choices}: We can grasp the impact of hyperparameters like learning rate and batch size on generalization. For instance, we understand why larger learning rates can be beneficial and why excessively large batch sizes can hinder generalization. This understanding has inspired novel techniques like Sharpness-Aware Minimization (SAM) \cite{foret2020sharpness} which explicitly promote generalization by seeking flatter minima in the loss landscape.
    \item \textbf{Weight Decay}: We can explain the regularization effects of weight decay and understand its various mechanisms for improving generalization \cite{zhang2018three}.
    \item \textbf{Double Descent and Over-parameterization}: We can understand the regularization effect of over-parameterization and how it helps mitigate overfitting in certain regimes \citep{hastie2022surprises, mei2022generalization, adlam2020understanding}.
    \item \textbf{Weight Sharing and Pooling}: Weight sharing in convolutional neural networks (CNNs) induces equivalences in the function space, and global average pooling further enforces invariance. These properties provably enhance generalization by reducing (regularizing) the complexity of the function class \citep{mei2021learning}.
   \item \textbf{Locality and Hierarchy}: Locality (e.g., local receptive fields) regularizes the network by prioritizing the learning of local interactions (simpler) before long-range ones \cite{misiakiewicz2022learning, favero2021locality}. Hierarchy, coupled with locality, allows models to learn complex functions in a balanced manner, capturing both local (higher-order interactions) and global (lower-order interactions) information \cite{xiao2022eigenspace}. These properties help explain why Convolutional Neural Networks (CNNs) often generalize better than Multi-Layer Perceptrons (MLPs), and why deeper CNNs often outperform shallower ones.

\end{itemize}

% \textbf{The Challenge of Scaling Laws}
However, when it comes to scaling, the guiding principles become less clear. One foundamental challenge is the existence of the scaling law crossover phenomenon. This phenomenon makes it unwise to optimize performance for any static scale, as such an approach may only be effective up to a certain point and fail to generalize to larger scales, i.e., it risks overfitting to a finite scale. Unfortunately, scaling law crossover is likely unavoidable in practice. Any observed scaling law represents a specific, and likely suboptimal, trajectory through a vast space of possible scaling strategies.

To illustrate this, consider training a sequence of transformers with increasing flop budgets $f\in\mathbb N$. We can approximate the flops as $f \approx 6\mathscr{ND}$, where $\mathscr{N}$ is the number of parameters and $\mathscr{D}$ is the number of data tokens. The loss, $\mathscr{L}$, depends on numerous factors including width ($D$), layers ($L$), $\mathscr{D}$, batch size ($B$), learning rate ($\eta$), weight decay ($\lambda$), and others. For simplicity, let's focus on these six, recognizing that $\mathscr{N} \approx 12D^2L$ for transformers with $F=4D$.

With the constraint $f = 6 \cdot 12 D^2L \cdot \mathscr{D}$, we have five free variables. A scaling rule, $\phi$, dictates how these variables scale with $f$:

\begin{align}
\mathcal {SR} =\{\phi: f\in \mathbb N\mapsto (D, L, \mathscr{D}, B, \eta, \lambda)\in \mathbb N^4 \times \mathbb R_+^2  \quad \text{with}\quad f = 72 D^2L \mathscr{D} \}
\end{align}

Each $\phi$ produces a scaling law curve within a 5-dimensional surface. Unless we identify the optimal scaling rule(s), $\phi^*$, that minimize $\mathscr{L}$ for all $f$:

\begin{equation*}
\mathscr{L}(\phi^*(f)) = \inf_{\phi\in\mathcal SR} \mathscr{L}(\phi(f)),
\end{equation*}

any given scaling law can likely be crossed by another. Since practical scaling rules are often heuristic and unlikely optimal, crossover is to be expected. For example:

\begin{itemize}
    \item We did not realize that Kaplan's scaling rule $\mathscr{D}\propto f^{0.27}$ \citep{kaplan2020scaling} was sub-optimal until the discovery of a better scaling rule $\mathscr{D}\propto f^{0.5}$ \citep{hoffmann2022training}. But how do we know that the Chinchilla scaling rule is optimal? Indeed, the exponent $\alpha=0.5$ is not universal and depends on the dataset in a complicated manner \citep{paquette2024}. In practice, a Chinchilla-type empirical analysis is often needed to determine the optimal scaling relationship between $\mathscr D$ and $\mathscr N$ for new datasets \citep{bi2024deepseek, dubey2024llama}.

\item Section \ref{sec:sub-optimal learning rate} shows that a constant learning rate scaling rule is sub-optimal, as scaling the learning rate with $2/D$ leads to better scaling \citep{yang2022tensor, everett2024scaling}. But how do we know there isn't an even better scaling rule?

\item Section \ref{sec: weight-decay suboptimal} demonstrates that a constant, independent weight decay is sub-optimal, since decaying weight decay with the learning rate leads to better scaling.  Again, how do we know there isn't a better approach?
\end{itemize}

Therefore, there is little hope of identifying truly optimal scaling rules $\phi^*$ in practice. It is more likely that we will gradually identify better practices and better methodologies for scaling. To do so effectively and reliably, we need guiding principles to navigate the complex, high-dimensional scaling space $\mathcal{SR}$\footnote{In practice, the dimensionality of the scaling space is much greater than what is assumed here.} and facilitate meaningful comparisons between different scaling strategies.

This leads to a central question in scaling:

\begin{center}
\textit{``What are the guiding principles for scaling that enable model comparison at scale?''}
\end{center}

\section{Limitation.} Machine learning is a rapidly evolving field, and our current understanding of scaling phenomena remains limited. What holds true today may not be valid in a few months. Notably, we mainly focus on reducing the loss in pretraining and our analysis assumes a ``skydiving'' regime, where data complexity significantly exceeds model complexity. This assumption may break in at least two settings. 
First, in post-training (e.g., instructional finetuning), the number of tokens are smaller than the number of parameters and reducing overfitting is important. We didn't dive into post-training in this paper. 
Second, 
as computational resources grow exponentially while high-quality data may plateau, it is possible that we will re-enter a U-shaped regime or even the second-descent regime (Fig.~\ref{fig:sky-diving}). In this case, traditional wisdom may regain relevance, and ``new wisdom'' may become outdated.  

\section*{Acknowledgement.} We express our gratitude to Atish Agarwala, Katie Everett, Ran Tian, Elliot Paquette and Yasaman Bahri for their insightful comments and invaluable suggestions. We also thank the members of the PAGI SoS team members Jaehoon Lee, Ben Adlam, Jeffrey Pennington, Roman Novak,  Izzeddin Gur, Alex Alemi, Jiri Hron, Gamaleldin Elsayed, Kelvin Xu, Peter J. Liu, Ian Fisher Mark Kurzeja, Abhishek Kumar, Justin Gilmer, Xinyang Geng, Jascha Sohl-dickstein, Mitchell Wortsman, Hugo Larochelle and PAGI Core for their  support throughout this project. Additionally, we acknowledge the valuable assistance of Gemini in refining the writing of this work. Finally, we thank the JAX \citep{jax2018github}, FLAX \citep{flax2020github}, Optax \citep{deepmind2020jax}, and NanoDO \citep{nanodo} teams for providing the essential infrastructure that made this research possible.

\bibliography{main}

\begin{thebibliography}{98}
\providecommand{\natexlab}[1]{#1}
\providecommand{\url}[1]{\texttt{#1}}
\expandafter\ifx\csname urlstyle\endcsname\relax
  \providecommand{\doi}[1]{doi: #1}\else
  \providecommand{\doi}{doi: \begingroup \urlstyle{rm}\Url}\fi

\bibitem[Achiam et~al.(2023)Achiam, Adler, Agarwal, Ahmad, Akkaya, Aleman, Almeida, Altenschmidt, Altman, Anadkat, et~al.]{achiam2023gpt}
Josh Achiam, Steven Adler, Sandhini Agarwal, Lama Ahmad, Ilge Akkaya, Florencia~Leoni Aleman, Diogo Almeida, Janko Altenschmidt, Sam Altman, Shyamal Anadkat, et~al.
\newblock Gpt-4 technical report.
\newblock \emph{arXiv preprint arXiv:2303.08774}, 2023.

\bibitem[Adlam \& Pennington(2020)Adlam and Pennington]{adlam2020understanding}
Ben Adlam and Jeffrey Pennington.
\newblock Understanding double descent requires a fine-grained bias-variance decomposition.
\newblock \emph{Advances in neural information processing systems}, 33:\penalty0 11022--11032, 2020.

\bibitem[Agarwala et~al.(2022)Agarwala, Pedregosa, and Pennington]{agarwala2022second}
Atish Agarwala, Fabian Pedregosa, and Jeffrey Pennington.
\newblock Second-order regression models exhibit progressive sharpening to the edge of stability.
\newblock \emph{arXiv preprint arXiv:2210.04860}, 2022.

\bibitem[Ainslie et~al.(2023)Ainslie, Lee-Thorp, de~Jong, Zemlyanskiy, Lebr{\'o}n, and Sanghai]{ainslie2023gqa}
Joshua Ainslie, James Lee-Thorp, Michiel de~Jong, Yury Zemlyanskiy, Federico Lebr{\'o}n, and Sumit Sanghai.
\newblock Gqa: Training generalized multi-query transformer models from multi-head checkpoints.
\newblock \emph{arXiv preprint arXiv:2305.13245}, 2023.

\bibitem[Andriushchenko et~al.(2023)Andriushchenko, D'Angelo, Varre, and Flammarion]{andriushchenko2023we}
Maksym Andriushchenko, Francesco D'Angelo, Aditya Varre, and Nicolas Flammarion.
\newblock Why do we need weight decay in modern deep learning?
\newblock \emph{arXiv preprint arXiv:2310.04415}, 2023.

\bibitem[Anil et~al.(2023)Anil, Borgeaud, Wu, Alayrac, Yu, Soricut, Schalkwyk, Dai, Hauth, Millican, et~al.]{anil2023gemini}
Rohan Anil, Sebastian Borgeaud, Yonghui Wu, Jean-Baptiste Alayrac, Jiahui Yu, Radu Soricut, Johan Schalkwyk, Andrew~M Dai, Anja Hauth, Katie Millican, et~al.
\newblock Gemini: A family of highly capable multimodal models.
\newblock \emph{arXiv preprint arXiv:2312.11805}, 1, 2023.

\bibitem[Bahri et~al.(2024)Bahri, Dyer, Kaplan, Lee, and Sharma]{bahri2024explaining}
Yasaman Bahri, Ethan Dyer, Jared Kaplan, Jaehoon Lee, and Utkarsh Sharma.
\newblock Explaining neural scaling laws.
\newblock \emph{Proceedings of the National Academy of Sciences}, 121\penalty0 (27):\penalty0 e2311878121, 2024.

\bibitem[Bartlett et~al.(2020)Bartlett, Long, Lugosi, and Tsigler]{bartlett2020benign}
Peter~L Bartlett, Philip~M Long, G{\'a}bor Lugosi, and Alexander Tsigler.
\newblock Benign overfitting in linear regression.
\newblock \emph{Proceedings of the National Academy of Sciences}, 117\penalty0 (48):\penalty0 30063--30070, 2020.

\bibitem[Beck et~al.(2024)Beck, P{\"o}ppel, Spanring, Auer, Prudnikova, Kopp, Klambauer, Brandstetter, and Hochreiter]{beck2024xlstm}
Maximilian Beck, Korbinian P{\"o}ppel, Markus Spanring, Andreas Auer, Oleksandra Prudnikova, Michael Kopp, G{\"u}nter Klambauer, Johannes Brandstetter, and Sepp Hochreiter.
\newblock xlstm: Extended long short-term memory.
\newblock \emph{arXiv preprint arXiv:2405.04517}, 2024.

\bibitem[Belkin et~al.(2019)Belkin, Hsu, Ma, and Mandal]{belkin2019reconciling}
Mikhail Belkin, Daniel Hsu, Siyuan Ma, and Soumik Mandal.
\newblock Reconciling modern machine-learning practice and the classical bias--variance trade-off.
\newblock \emph{Proceedings of the National Academy of Sciences}, 116\penalty0 (32):\penalty0 15849--15854, 2019.

\bibitem[bench authors(2023)]{srivastava2023beyond}
BIG bench authors.
\newblock Beyond the imitation game: Quantifying and extrapolating the capabilities of language models.
\newblock \emph{Transactions on Machine Learning Research}, 2023.
\newblock ISSN 2835-8856.
\newblock URL \url{https://openreview.net/forum?id=uyTL5Bvosj}.

\bibitem[Besiroglu et~al.(2024)Besiroglu, Erdil, Barnett, and You]{besiroglu2024chinchilla}
Tamay Besiroglu, Ege Erdil, Matthew Barnett, and Josh You.
\newblock Chinchilla scaling: A replication attempt.
\newblock \emph{arXiv preprint arXiv:2404.10102}, 2024.

\bibitem[Bi et~al.(2024)Bi, Chen, Chen, Chen, Dai, Deng, Ding, Dong, Du, Fu, et~al.]{bi2024deepseek}
Xiao Bi, Deli Chen, Guanting Chen, Shanhuang Chen, Damai Dai, Chengqi Deng, Honghui Ding, Kai Dong, Qiushi Du, Zhe Fu, et~al.
\newblock Deepseek llm: Scaling open-source language models with longtermism.
\newblock \emph{arXiv preprint arXiv:2401.02954}, 2024.

\bibitem[Blake et~al.(2024)Blake, Eichenberg, Dean, Balles, Prince, Deiseroth, Cruz-Salinas, Luschi, Weinbach, and Orr]{blake2024u}
Charlie Blake, Constantin Eichenberg, Josef Dean, Lukas Balles, Luke~Y Prince, Bj{\"o}rn Deiseroth, Andres~Felipe Cruz-Salinas, Carlo Luschi, Samuel Weinbach, and Douglas Orr.
\newblock u-$\mu$p: The unit-scaled maximal update parametrization.
\newblock \emph{arXiv preprint arXiv:2407.17465}, 2024.

\bibitem[Bordelon et~al.(2023)Bordelon, Noci, Li, Hanin, and Pehlevan]{bordelon2023depthwise}
Blake Bordelon, Lorenzo Noci, Mufan~Bill Li, Boris Hanin, and Cengiz Pehlevan.
\newblock Depthwise hyperparameter transfer in residual networks: Dynamics and scaling limit.
\newblock \emph{arXiv preprint arXiv:2309.16620}, 2023.

\bibitem[Bordelon et~al.(2024)Bordelon, Atanasov, and Pehlevan]{bordelon2024dynamical}
Blake Bordelon, Alexander Atanasov, and Cengiz Pehlevan.
\newblock A dynamical model of neural scaling laws.
\newblock \emph{arXiv preprint arXiv:2402.01092}, 2024.

\bibitem[Botev et~al.(2024)Botev, De, Smith, Fernando, Muraru, Haroun, Berrada, Pascanu, Sessa, Dadashi, et~al.]{botev2024recurrentgemma}
Aleksandar Botev, Soham De, Samuel~L Smith, Anushan Fernando, George-Cristian Muraru, Ruba Haroun, Leonard Berrada, Razvan Pascanu, Pier~Giuseppe Sessa, Robert Dadashi, et~al.
\newblock Recurrentgemma: Moving past transformers for efficient open language models.
\newblock \emph{arXiv preprint arXiv:2404.07839}, 2024.

\bibitem[Bradbury et~al.(2018)Bradbury, Frostig, Hawkins, Johnson, Leary, Maclaurin, Necula, Paszke, Vander{P}las, Wanderman-{M}ilne, and Zhang]{jax2018github}
James Bradbury, Roy Frostig, Peter Hawkins, Matthew~James Johnson, Chris Leary, Dougal Maclaurin, George Necula, Adam Paszke, Jake Vander{P}las, Skye Wanderman-{M}ilne, and Qiao Zhang.
\newblock {JAX}: composable transformations of {P}ython+{N}um{P}y programs, 2018.
\newblock URL \url{http://github.com/google/jax}.

\bibitem[Brown et~al.(2020)Brown, Mann, Ryder, Subbiah, Kaplan, Dhariwal, Neelakantan, Shyam, Sastry, Askell, et~al.]{brown2020language}
Tom Brown, Benjamin Mann, Nick Ryder, Melanie Subbiah, Jared~D Kaplan, Prafulla Dhariwal, Arvind Neelakantan, Pranav Shyam, Girish Sastry, Amanda Askell, et~al.
\newblock Language models are few-shot learners.
\newblock \emph{Advances in neural information processing systems}, 33:\penalty0 1877--1901, 2020.

\bibitem[Chen et~al.(2024)Chen, Liang, Huang, Real, Wang, Pham, Dong, Luong, Hsieh, Lu, et~al.]{chen2024symbolic}
Xiangning Chen, Chen Liang, Da~Huang, Esteban Real, Kaiyuan Wang, Hieu Pham, Xuanyi Dong, Thang Luong, Cho-Jui Hsieh, Yifeng Lu, et~al.
\newblock Symbolic discovery of optimization algorithms.
\newblock \emph{Advances in neural information processing systems}, 36, 2024.

\bibitem[Chizat et~al.(2019)Chizat, Oyallon, and Bach]{chizat2019lazy}
Lenaic Chizat, Edouard Oyallon, and Francis Bach.
\newblock On lazy training in differentiable programming.
\newblock \emph{Advances in neural information processing systems}, 32, 2019.

\bibitem[Chowdhery et~al.(2023)Chowdhery, Narang, Devlin, Bosma, Mishra, Roberts, Barham, Chung, Sutton, Gehrmann, et~al.]{chowdhery2023palm}
Aakanksha Chowdhery, Sharan Narang, Jacob Devlin, Maarten Bosma, Gaurav Mishra, Adam Roberts, Paul Barham, Hyung~Won Chung, Charles Sutton, Sebastian Gehrmann, et~al.
\newblock Palm: Scaling language modeling with pathways.
\newblock \emph{Journal of Machine Learning Research}, 24\penalty0 (240):\penalty0 1--113, 2023.

\bibitem[Cohen et~al.(2020)Cohen, Kaur, Li, Kolter, and Talwalkar]{cohen2020gradient}
Jeremy Cohen, Simran Kaur, Yuanzhi Li, J~Zico Kolter, and Ameet Talwalkar.
\newblock Gradient descent on neural networks typically occurs at the edge of stability.
\newblock In \emph{International Conference on Learning Representations}, 2020.

\bibitem[Cohen et~al.(2022)Cohen, Ghorbani, Krishnan, Agarwal, Medapati, Badura, Suo, Cardoze, Nado, Dahl, et~al.]{cohen2022adaptive}
Jeremy~M Cohen, Behrooz Ghorbani, Shankar Krishnan, Naman Agarwal, Sourabh Medapati, Michal Badura, Daniel Suo, David Cardoze, Zachary Nado, George~E Dahl, et~al.
\newblock Adaptive gradient methods at the edge of stability.
\newblock \emph{arXiv preprint arXiv:2207.14484}, 2022.

\bibitem[Damian et~al.(2022)Damian, Nichani, and Lee]{damian2022self}
Alex Damian, Eshaan Nichani, and Jason~D Lee.
\newblock Self-stabilization: The implicit bias of gradient descent at the edge of stability.
\newblock \emph{arXiv preprint arXiv:2209.15594}, 2022.

\bibitem[DeepMind et~al.(2020)DeepMind, Babuschkin, Baumli, Bell, Bhupatiraju, Bruce, Buchlovsky, Budden, Cai, Clark, Danihelka, Dedieu, Fantacci, Godwin, Jones, Hemsley, Hennigan, Hessel, Hou, Kapturowski, Keck, Kemaev, King, Kunesch, Martens, Merzic, Mikulik, Norman, Papamakarios, Quan, Ring, Ruiz, Sanchez, Sartran, Schneider, Sezener, Spencer, Srinivasan, Stanojevi\'{c}, Stokowiec, Wang, Zhou, and Viola]{deepmind2020jax}
DeepMind, Igor Babuschkin, Kate Baumli, Alison Bell, Surya Bhupatiraju, Jake Bruce, Peter Buchlovsky, David Budden, Trevor Cai, Aidan Clark, Ivo Danihelka, Antoine Dedieu, Claudio Fantacci, Jonathan Godwin, Chris Jones, Ross Hemsley, Tom Hennigan, Matteo Hessel, Shaobo Hou, Steven Kapturowski, Thomas Keck, Iurii Kemaev, Michael King, Markus Kunesch, Lena Martens, Hamza Merzic, Vladimir Mikulik, Tamara Norman, George Papamakarios, John Quan, Roman Ring, Francisco Ruiz, Alvaro Sanchez, Laurent Sartran, Rosalia Schneider, Eren Sezener, Stephen Spencer, Srivatsan Srinivasan, Milo\v{s} Stanojevi\'{c}, Wojciech Stokowiec, Luyu Wang, Guangyao Zhou, and Fabio Viola.
\newblock The {D}eep{M}ind {JAX} {E}cosystem, 2020.
\newblock URL \url{http://github.com/google-deepmind}.

\bibitem[DeepSeek-AI et~al.(2024)DeepSeek-AI, Liu, Feng, Wang, Wang, Liu, Zhao, Dengr, Ruan, Dai, Guo, Yang, Chen, Ji, Li, Lin, Luo, Hao, Chen, Li, Zhang, Xu, Yang, Zhang, Ding, Xin, Gao, Li, Qu, Cai, Liang, Guo, Ni, Li, Chen, Yuan, Qiu, Song, Dong, Gao, Guan, Wang, Zhang, Xu, Xia, Zhao, Zhang, Li, Wang, Zhang, Zhang, Tang, Li, Tian, Huang, Wang, Zhang, Zhu, Chen, Du, Chen, Jin, Ge, Pan, Xu, Chen, Li, Lu, Zhou, Chen, Wu, Ye, Ma, Wang, Zhou, Yu, Zhou, Zheng, Wang, Pei, Yuan, Sun, Xiao, Zeng, An, Liu, Liang, Gao, Zhang, Li, Jin, Wang, Bi, Liu, Wang, Shen, Chen, Chen, Nie, Sun, Wang, Liu, Xie, Yu, Song, Zhou, Yang, Lu, Su, Wu, Li, Wei, Zhu, Xu, Huang, Li, Zhao, Sun, Li, Wang, Zheng, Zhang, Xiong, Zhao, He, Tang, Piao, Dong, Tan, Liu, Wang, Guo, Zhu, Wang, Zou, Zha, Ma, Yan, You, Liu, Ren, Ren, Sha, Fu, Huang, Zhang, Xie, Hao, Shao, Wen, Xu, Zhang, Li, Wang, Gu, Li, and Xie]{deepseekai2024deepseekv2strongeconomicalefficient}
DeepSeek-AI, Aixin Liu, Bei Feng, Bin Wang, Bingxuan Wang, Bo~Liu, Chenggang Zhao, Chengqi Dengr, Chong Ruan, Damai Dai, Daya Guo, Dejian Yang, Deli Chen, Dongjie Ji, Erhang Li, Fangyun Lin, Fuli Luo, Guangbo Hao, Guanting Chen, Guowei Li, H.~Zhang, Hanwei Xu, Hao Yang, Haowei Zhang, Honghui Ding, Huajian Xin, Huazuo Gao, Hui Li, Hui Qu, J.~L. Cai, Jian Liang, Jianzhong Guo, Jiaqi Ni, Jiashi Li, Jin Chen, Jingyang Yuan, Junjie Qiu, Junxiao Song, Kai Dong, Kaige Gao, Kang Guan, Lean Wang, Lecong Zhang, Lei Xu, Leyi Xia, Liang Zhao, Liyue Zhang, Meng Li, Miaojun Wang, Mingchuan Zhang, Minghua Zhang, Minghui Tang, Mingming Li, Ning Tian, Panpan Huang, Peiyi Wang, Peng Zhang, Qihao Zhu, Qinyu Chen, Qiushi Du, R.~J. Chen, R.~L. Jin, Ruiqi Ge, Ruizhe Pan, Runxin Xu, Ruyi Chen, S.~S. Li, Shanghao Lu, Shangyan Zhou, Shanhuang Chen, Shaoqing Wu, Shengfeng Ye, Shirong Ma, Shiyu Wang, Shuang Zhou, Shuiping Yu, Shunfeng Zhou, Size Zheng, T.~Wang, Tian Pei, Tian Yuan, Tianyu Sun, W.~L. Xiao, Wangding Zeng, Wei An, Wen
  Liu, Wenfeng Liang, Wenjun Gao, Wentao Zhang, X.~Q. Li, Xiangyue Jin, Xianzu Wang, Xiao Bi, Xiaodong Liu, Xiaohan Wang, Xiaojin Shen, Xiaokang Chen, Xiaosha Chen, Xiaotao Nie, Xiaowen Sun, Xiaoxiang Wang, Xin Liu, Xin Xie, Xingkai Yu, Xinnan Song, Xinyi Zhou, Xinyu Yang, Xuan Lu, Xuecheng Su, Y.~Wu, Y.~K. Li, Y.~X. Wei, Y.~X. Zhu, Yanhong Xu, Yanping Huang, Yao Li, Yao Zhao, Yaofeng Sun, Yaohui Li, Yaohui Wang, Yi~Zheng, Yichao Zhang, Yiliang Xiong, Yilong Zhao, Ying He, Ying Tang, Yishi Piao, Yixin Dong, Yixuan Tan, Yiyuan Liu, Yongji Wang, Yongqiang Guo, Yuchen Zhu, Yuduan Wang, Yuheng Zou, Yukun Zha, Yunxian Ma, Yuting Yan, Yuxiang You, Yuxuan Liu, Z.~Z. Ren, Zehui Ren, Zhangli Sha, Zhe Fu, Zhen Huang, Zhen Zhang, Zhenda Xie, Zhewen Hao, Zhihong Shao, Zhiniu Wen, Zhipeng Xu, Zhongyu Zhang, Zhuoshu Li, Zihan Wang, Zihui Gu, Zilin Li, and Ziwei Xie.
\newblock Deepseek-v2: A strong, economical, and efficient mixture-of-experts language model, 2024.
\newblock URL \url{https://arxiv.org/abs/2405.04434}.

\bibitem[Defazio et~al.(2024)Defazio, Mehta, Mishchenko, Khaled, Cutkosky, et~al.]{defazio2024road}
Aaron Defazio, Harsh Mehta, Konstantin Mishchenko, Ahmed Khaled, Ashok Cutkosky, et~al.
\newblock The road less scheduled.
\newblock \emph{arXiv preprint arXiv:2405.15682}, 2024.

\bibitem[Dehghani et~al.(2023)Dehghani, Djolonga, Mustafa, Padlewski, Heek, Gilmer, Steiner, Caron, Geirhos, Alabdulmohsin, Jenatton, Beyer, Tschannen, Arnab, Wang, Riquelme~Ruiz, Minderer, Puigcerver, Evci, Kumar, Steenkiste, Elsayed, Mahendran, Yu, Oliver, Huot, Bastings, Collier, Gritsenko, Birodkar, Vasconcelos, Tay, Mensink, Kolesnikov, Pavetic, Tran, Kipf, Lucic, Zhai, Keysers, Harmsen, and Houlsby]{pmlr-v202-dehghani23a}
Mostafa Dehghani, Josip Djolonga, Basil Mustafa, Piotr Padlewski, Jonathan Heek, Justin Gilmer, Andreas~Peter Steiner, Mathilde Caron, Robert Geirhos, Ibrahim Alabdulmohsin, Rodolphe Jenatton, Lucas Beyer, Michael Tschannen, Anurag Arnab, Xiao Wang, Carlos Riquelme~Ruiz, Matthias Minderer, Joan Puigcerver, Utku Evci, Manoj Kumar, Sjoerd~Van Steenkiste, Gamaleldin~Fathy Elsayed, Aravindh Mahendran, Fisher Yu, Avital Oliver, Fantine Huot, Jasmijn Bastings, Mark Collier, Alexey~A. Gritsenko, Vighnesh Birodkar, Cristina~Nader Vasconcelos, Yi~Tay, Thomas Mensink, Alexander Kolesnikov, Filip Pavetic, Dustin Tran, Thomas Kipf, Mario Lucic, Xiaohua Zhai, Daniel Keysers, Jeremiah~J. Harmsen, and Neil Houlsby.
\newblock Scaling vision transformers to 22 billion parameters.
\newblock In Andreas Krause, Emma Brunskill, Kyunghyun Cho, Barbara Engelhardt, Sivan Sabato, and Jonathan Scarlett (eds.), \emph{Proceedings of the 40th International Conference on Machine Learning}, volume 202 of \emph{Proceedings of Machine Learning Research}, pp.\  7480--7512. PMLR, 23--29 Jul 2023.
\newblock URL \url{https://proceedings.mlr.press/v202/dehghani23a.html}.

\bibitem[Devlin et~al.(2018)Devlin, Chang, Lee, and Toutanova]{devlin2018bert}
Jacob Devlin, Ming-Wei Chang, Kenton Lee, and Kristina Toutanova.
\newblock Bert: Pre-training of deep bidirectional transformers for language understanding.
\newblock \emph{arXiv preprint arXiv:1810.04805}, 2018.

\bibitem[Dinh et~al.(2017)Dinh, Pascanu, Bengio, and Bengio]{dinh2017sharp}
Laurent Dinh, Razvan Pascanu, Samy Bengio, and Yoshua Bengio.
\newblock Sharp minima can generalize for deep nets.
\newblock In \emph{International Conference on Machine Learning}, pp.\  1019--1028. PMLR, 2017.

\bibitem[Dosovitskiy et~al.(2020)Dosovitskiy, Beyer, Kolesnikov, Weissenborn, Zhai, Unterthiner, Dehghani, Minderer, Heigold, Gelly, et~al.]{dosovitskiy2020image}
Alexey Dosovitskiy, Lucas Beyer, Alexander Kolesnikov, Dirk Weissenborn, Xiaohua Zhai, Thomas Unterthiner, Mostafa Dehghani, Matthias Minderer, Georg Heigold, Sylvain Gelly, et~al.
\newblock An image is worth 16x16 words: Transformers for image recognition at scale.
\newblock \emph{arXiv preprint arXiv:2010.11929}, 2020.

\bibitem[Dubey et~al.(2024)Dubey, Jauhri, Pandey, Kadian, Al-Dahle, Letman, Mathur, Schelten, Yang, Fan, et~al.]{dubey2024llama}
Abhimanyu Dubey, Abhinav Jauhri, Abhinav Pandey, Abhishek Kadian, Ahmad Al-Dahle, Aiesha Letman, Akhil Mathur, Alan Schelten, Amy Yang, Angela Fan, et~al.
\newblock The llama 3 herd of models.
\newblock \emph{arXiv preprint arXiv:2407.21783}, 2024.

\bibitem[Everett et~al.(2024)Everett, Xiao, Wortsman, Alemi, Novak, Liu, Gur, Sohl-Dickstein, Kaelbling, Lee, and Pennington]{everett2024scaling}
Katie~E Everett, Lechao Xiao, Mitchell Wortsman, Alexander~A Alemi, Roman Novak, Peter~J Liu, Izzeddin Gur, Jascha Sohl-Dickstein, Leslie~Pack Kaelbling, Jaehoon Lee, and Jeffrey Pennington.
\newblock Scaling exponents across parameterizations and optimizers.
\newblock In \emph{Forty-first International Conference on Machine Learning}, 2024.
\newblock URL \url{https://openreview.net/forum?id=0ksNeD1SJT}.

\bibitem[Favero et~al.(2021)Favero, Cagnetta, and Wyart]{favero2021locality}
Alessandro Favero, Francesco Cagnetta, and Matthieu Wyart.
\newblock Locality defeats the curse of dimensionality in convolutional teacher-student scenarios.
\newblock \emph{Advances in Neural Information Processing Systems}, 34:\penalty0 9456--9467, 2021.

\bibitem[Foret et~al.(2020)Foret, Kleiner, Mobahi, and Neyshabur]{foret2020sharpness}
Pierre Foret, Ariel Kleiner, Hossein Mobahi, and Behnam Neyshabur.
\newblock Sharpness-aware minimization for efficiently improving generalization.
\newblock \emph{arXiv preprint arXiv:2010.01412}, 2020.

\bibitem[Gilmer~J. \& J.(2023)Gilmer~J. and J.]{Gilmer2023}
Schioppa~A. Gilmer~J. and Cohen J.
\newblock Intriguing properties of transformer training instabilities.
\newblock \emph{To appear.}, 2023.

\bibitem[Goh(2017)]{goh2017why}
Gabriel Goh.
\newblock Why momentum really works.
\newblock \emph{Distill}, 2017.
\newblock \doi{10.23915/distill.00006}.
\newblock URL \url{http://distill.pub/2017/momentum}.

\bibitem[Gu \& Dao(2023)Gu and Dao]{gu2023mamba}
Albert Gu and Tri Dao.
\newblock Mamba: Linear-time sequence modeling with selective state spaces.
\newblock \emph{arXiv preprint arXiv:2312.00752}, 2023.

\bibitem[Gupta et~al.(2018)Gupta, Koren, and Singer]{gupta2018shampoo}
Vineet Gupta, Tomer Koren, and Yoram Singer.
\newblock Shampoo: Preconditioned stochastic tensor optimization.
\newblock In \emph{International Conference on Machine Learning}, pp.\  1842--1850. PMLR, 2018.

\bibitem[Hastie et~al.(2009)Hastie, Tibshirani, Friedman, and Friedman]{hastie2009elements}
Trevor Hastie, Robert Tibshirani, Jerome~H Friedman, and Jerome~H Friedman.
\newblock \emph{The elements of statistical learning: data mining, inference, and prediction}, volume~2.
\newblock Springer, 2009.

\bibitem[Hastie et~al.(2022)Hastie, Montanari, Rosset, and Tibshirani]{hastie2022surprises}
Trevor Hastie, Andrea Montanari, Saharon Rosset, and Ryan~J Tibshirani.
\newblock Surprises in high-dimensional ridgeless least squares interpolation.
\newblock \emph{Annals of statistics}, 50\penalty0 (2):\penalty0 949, 2022.

\bibitem[He et~al.(2016)He, Zhang, Ren, and Sun]{he2016deep}
Kaiming He, Xiangyu Zhang, Shaoqing Ren, and Jian Sun.
\newblock Deep residual learning for image recognition.
\newblock In \emph{Proceedings of the IEEE conference on computer vision and pattern recognition}, pp.\  770--778, 2016.

\bibitem[Heek et~al.(2023)Heek, Levskaya, Oliver, Ritter, Rondepierre, Steiner, and van {Z}ee]{flax2020github}
Jonathan Heek, Anselm Levskaya, Avital Oliver, Marvin Ritter, Bertrand Rondepierre, Andreas Steiner, and Marc van {Z}ee.
\newblock {F}lax: A neural network library and ecosystem for {JAX}, 2023.
\newblock URL \url{http://github.com/google/flax}.

\bibitem[Hendrycks \& Gimpel(2016)Hendrycks and Gimpel]{hendrycks2016gaussian}
Dan Hendrycks and Kevin Gimpel.
\newblock Gaussian error linear units (gelus).
\newblock \emph{arXiv preprint arXiv:1606.08415}, 2016.

\bibitem[Hestness et~al.(2017)Hestness, Narang, Ardalani, Diamos, Jun, Kianinejad, Patwary, Yang, and Zhou]{hestness2017deep}
Joel Hestness, Sharan Narang, Newsha Ardalani, Gregory Diamos, Heewoo Jun, Hassan Kianinejad, Md~Mostofa~Ali Patwary, Yang Yang, and Yanqi Zhou.
\newblock Deep learning scaling is predictable, empirically.
\newblock \emph{arXiv preprint arXiv:1712.00409}, 2017.

\bibitem[Hochreiter \& Schmidhuber(1997)Hochreiter and Schmidhuber]{hochreiter1997flat}
Sepp Hochreiter and J{\"u}rgen Schmidhuber.
\newblock Flat minima.
\newblock \emph{Neural computation}, 9\penalty0 (1):\penalty0 1--42, 1997.

\bibitem[Hoffmann et~al.(2022)Hoffmann, Borgeaud, Mensch, Buchatskaya, Cai, Rutherford, Casas, Hendricks, Welbl, Clark, et~al.]{hoffmann2022training}
Jordan Hoffmann, Sebastian Borgeaud, Arthur Mensch, Elena Buchatskaya, Trevor Cai, Eliza Rutherford, Diego de~Las Casas, Lisa~Anne Hendricks, Johannes Welbl, Aidan Clark, et~al.
\newblock Training compute-optimal large language models.
\newblock \emph{arXiv preprint arXiv:2203.15556}, 2022.

\bibitem[Jacot et~al.(2018)Jacot, Gabriel, and Hongler]{jacot2018neural}
Arthur Jacot, Franck Gabriel, and Cl{\'e}ment Hongler.
\newblock Neural tangent kernel: Convergence and generalization in neural networks.
\newblock \emph{Advances in neural information processing systems}, 31, 2018.

\bibitem[Kaplan et~al.(2020)Kaplan, McCandlish, Henighan, Brown, Chess, Child, Gray, Radford, Wu, and Amodei]{kaplan2020scaling}
Jared Kaplan, Sam McCandlish, Tom Henighan, Tom~B Brown, Benjamin Chess, Rewon Child, Scott Gray, Alec Radford, Jeffrey Wu, and Dario Amodei.
\newblock Scaling laws for neural language models.
\newblock \emph{arXiv preprint arXiv:2001.08361}, 2020.

\bibitem[Karkada(2024)]{karkada2024lazy}
Dhruva Karkada.
\newblock The lazy (ntk) and rich ($\mu$-p) regimes: a gentle tutorial.
\newblock \emph{arXiv preprint arXiv:2404.19719}, 2024.

\bibitem[Keskar et~al.(2016)Keskar, Mudigere, Nocedal, Smelyanskiy, and Tang]{keskar2016large}
Nitish~Shirish Keskar, Dheevatsa Mudigere, Jorge Nocedal, Mikhail Smelyanskiy, and Ping Tak~Peter Tang.
\newblock On large-batch training for deep learning: Generalization gap and sharp minima.
\newblock \emph{arXiv preprint arXiv:1609.04836}, 2016.

\bibitem[Kingma \& Ba(2014)Kingma and Ba]{kingma2014adam}
Diederik~P Kingma and Jimmy Ba.
\newblock Adam: A method for stochastic optimization.
\newblock \emph{arXiv preprint arXiv:1412.6980}, 2014.

\bibitem[Krizhevsky et~al.(2012)Krizhevsky, Sutskever, and Hinton]{krizhevsky2012imagenet}
Alex Krizhevsky, Ilya Sutskever, and Geoffrey~E Hinton.
\newblock Imagenet classification with deep convolutional neural networks.
\newblock \emph{Advances in neural information processing systems}, 25, 2012.

\bibitem[Lee et~al.(2019)Lee, Xiao, Schoenholz, Bahri, Novak, Sohl-Dickstein, and Pennington]{lee2019wide}
Jaehoon Lee, Lechao Xiao, Samuel Schoenholz, Yasaman Bahri, Roman Novak, Jascha Sohl-Dickstein, and Jeffrey Pennington.
\newblock Wide neural networks of any depth evolve as linear models under gradient descent.
\newblock \emph{Advances in neural information processing systems}, 32, 2019.

\bibitem[Lee et~al.(2020)Lee, Schoenholz, Pennington, Adlam, Xiao, Novak, and Sohl-Dickstein]{lee2020finite}
Jaehoon Lee, Samuel Schoenholz, Jeffrey Pennington, Ben Adlam, Lechao Xiao, Roman Novak, and Jascha Sohl-Dickstein.
\newblock Finite versus infinite neural networks: an empirical study.
\newblock \emph{Advances in Neural Information Processing Systems}, 33:\penalty0 15156--15172, 2020.

\bibitem[Lewkowycz et~al.(2020)Lewkowycz, Bahri, Dyer, Sohl-Dickstein, and Gur-Ari]{lewkowycz2020large}
Aitor Lewkowycz, Yasaman Bahri, Ethan Dyer, Jascha Sohl-Dickstein, and Guy Gur-Ari.
\newblock The large learning rate phase of deep learning: the catapult mechanism.
\newblock \emph{arXiv preprint arXiv:2003.02218}, 2020.

\bibitem[Li et~al.(2019)Li, Wei, and Ma]{li2019towards}
Yuanzhi Li, Colin Wei, and Tengyu Ma.
\newblock Towards explaining the regularization effect of initial large learning rate in training neural networks.
\newblock \emph{Advances in neural information processing systems}, 32, 2019.

\bibitem[Lin et~al.(2024)Lin, Wu, Kakade, Bartlett, and Lee]{lin2024scaling}
Licong Lin, Jingfeng Wu, Sham~M Kakade, Peter~L Bartlett, and Jason~D Lee.
\newblock Scaling laws in linear regression: Compute, parameters, and data.
\newblock \emph{arXiv preprint arXiv:2406.08466}, 2024.

\bibitem[Lingle(2024)]{lingle2024large}
Lucas Lingle.
\newblock A large-scale exploration of $\mu$-transfer.
\newblock \emph{arXiv preprint arXiv:2404.05728}, 2024.

\bibitem[Liu et~al.(2020)Liu, Liu, Gao, Chen, and Han]{liu2020understanding}
Liyuan Liu, Xiaodong Liu, Jianfeng Gao, Weizhu Chen, and Jiawei Han.
\newblock Understanding the difficulty of training transformers.
\newblock \emph{arXiv preprint arXiv:2004.08249}, 2020.

\bibitem[Liu et~al.(2024)Liu, Novak, Lee, Wortsman, Xiao, Everett, Alemi, Kurzeja, Marcenac, Gur, Kornblith, Xu, Elsayed, Fischer, Pennington, Adlam, and Dickstein]{nanodo}
Peter~J. Liu, Roman Novak, Jaehoon Lee, Mitchell Wortsman, Lechao Xiao, Katie Everett, Alexander~A. Alemi, Mark Kurzeja, Pierre Marcenac, Izzeddin Gur, Simon Kornblith, Kelvin Xu, Gamaleldin Elsayed, Ian Fischer, Jeffrey Pennington, Ben Adlam, and Jascha-Sohl Dickstein.
\newblock Nanodo: A minimal transformer decoder-only language model implementation in {JAX}., 2024.
\newblock URL \url{http://github.com/google-deepmind/nanodo}.

\bibitem[Loshchilov \& Hutter(2017)Loshchilov and Hutter]{loshchilov2017decoupled}
Ilya Loshchilov and Frank Hutter.
\newblock Decoupled weight decay regularization.
\newblock \emph{arXiv preprint arXiv:1711.05101}, 2017.

\bibitem[McCandlish et~al.(2018)McCandlish, Kaplan, Amodei, and Team]{mccandlish2018empirical}
Sam McCandlish, Jared Kaplan, Dario Amodei, and OpenAI~Dota Team.
\newblock An empirical model of large-batch training.
\newblock \emph{arXiv preprint arXiv:1812.06162}, 2018.

\bibitem[Mei \& Montanari(2022)Mei and Montanari]{mei2022generalization}
Song Mei and Andrea Montanari.
\newblock The generalization error of random features regression: Precise asymptotics and the double descent curve.
\newblock \emph{Communications on Pure and Applied Mathematics}, 75\penalty0 (4):\penalty0 667--766, 2022.

\bibitem[Mei et~al.(2021)Mei, Misiakiewicz, and Montanari]{mei2021learning}
Song Mei, Theodor Misiakiewicz, and Andrea Montanari.
\newblock Learning with invariances in random features and kernel models.
\newblock In \emph{Conference on Learning Theory}, pp.\  3351--3418. PMLR, 2021.

\bibitem[Misiakiewicz \& Mei(2022)Misiakiewicz and Mei]{misiakiewicz2022learning}
Theodor Misiakiewicz and Song Mei.
\newblock Learning with convolution and pooling operations in kernel methods.
\newblock \emph{Advances in Neural Information Processing Systems}, 35:\penalty0 29014--29025, 2022.

\bibitem[Molybog et~al.(2023)Molybog, Albert, Chen, DeVito, Esiobu, Goyal, Koura, Narang, Poulton, Silva, et~al.]{molybog2023theory}
Igor Molybog, Peter Albert, Moya Chen, Zachary DeVito, David Esiobu, Naman Goyal, Punit~Singh Koura, Sharan Narang, Andrew Poulton, Ruan Silva, et~al.
\newblock A theory on adam instability in large-scale machine learning.
\newblock \emph{arXiv preprint arXiv:2304.09871}, 2023.

\bibitem[Neyshabur(2017)]{neyshabur2017implicit}
Behnam Neyshabur.
\newblock Implicit regularization in deep learning.
\newblock \emph{arXiv preprint arXiv:1709.01953}, 2017.

\bibitem[Neyshabur et~al.(2018)Neyshabur, Li, Bhojanapalli, LeCun, and Srebro]{neyshabur2018towards}
Behnam Neyshabur, Zhiyuan Li, Srinadh Bhojanapalli, Yann LeCun, and Nathan Srebro.
\newblock Towards understanding the role of over-parametrization in generalization of neural networks.
\newblock \emph{arXiv preprint arXiv:1805.12076}, 2018.

\bibitem[OpenAI et~al.(2023)]{openai2023gpt}
R~OpenAI et~al.
\newblock Gpt-4 technical report.
\newblock \emph{ArXiv}, 2303:\penalty0 08774, 2023.

\bibitem[Paquette et~al.(2024)Paquette, Paquette, Xiao, and Pennington]{paquette2024}
Elliot Paquette, Courtney Paquette, Lechao Xiao, and Jeffrey Pennington.
\newblock 4+ 3 phases of compute-optimal neural scaling laws.
\newblock \emph{arXiv preprint arXiv:2405.15074}, 2024.

\bibitem[Penedo et~al.(2024)Penedo, Kydlíček, allal, Lozhkov, Mitchell, Raffel, Werra, and Wolf]{penedo2024finewebdatasetsdecantingweb}
Guilherme Penedo, Hynek Kydlíček, Loubna~Ben allal, Anton Lozhkov, Margaret Mitchell, Colin Raffel, Leandro~Von Werra, and Thomas Wolf.
\newblock The fineweb datasets: Decanting the web for the finest text data at scale, 2024.
\newblock URL \url{https://arxiv.org/abs/2406.17557}.

\bibitem[Porian et~al.(2024)Porian, Wortsman, Jitsev, Schmidt, and Carmon]{porian2024resolving}
Tomer Porian, Mitchell Wortsman, Jenia Jitsev, Ludwig Schmidt, and Yair Carmon.
\newblock Resolving discrepancies in compute-optimal scaling of language models.
\newblock \emph{arXiv preprint arXiv:2406.19146}, 2024.

\bibitem[Radford et~al.(2019)Radford, Wu, Child, Luan, Amodei, Sutskever, et~al.]{radford2019language}
Alec Radford, Jeffrey Wu, Rewon Child, David Luan, Dario Amodei, Ilya Sutskever, et~al.
\newblock Language models are unsupervised multitask learners.
\newblock \emph{OpenAI blog}, 1\penalty0 (8):\penalty0 9, 2019.

\bibitem[Raffel et~al.(2020)Raffel, Shazeer, Roberts, Lee, Narang, Matena, Zhou, Li, and Liu]{raffel2020exploring}
Colin Raffel, Noam Shazeer, Adam Roberts, Katherine Lee, Sharan Narang, Michael Matena, Yanqi Zhou, Wei Li, and Peter~J Liu.
\newblock Exploring the limits of transfer learning with a unified text-to-text transformer.
\newblock \emph{Journal of machine learning research}, 21\penalty0 (140):\penalty0 1--67, 2020.

\bibitem[Shallue et~al.(2019)Shallue, Lee, Antognini, Sohl-Dickstein, Frostig, and Dahl]{shallue2019measuring}
Christopher~J Shallue, Jaehoon Lee, Joseph Antognini, Jascha Sohl-Dickstein, Roy Frostig, and George~E Dahl.
\newblock Measuring the effects of data parallelism on neural network training.
\newblock \emph{Journal of Machine Learning Research}, 20\penalty0 (112):\penalty0 1--49, 2019.

\bibitem[Shazeer(2019)]{shazeer2019fast}
Noam Shazeer.
\newblock Fast transformer decoding: One write-head is all you need.
\newblock \emph{arXiv preprint arXiv:1911.02150}, 2019.

\bibitem[Shazeer(2020)]{shazeer2020glu}
Noam Shazeer.
\newblock Glu variants improve transformer.
\newblock \emph{arXiv preprint arXiv:2002.05202}, 2020.

\bibitem[Smith \& Le(2017)Smith and Le]{smith2017bayesian}
Samuel~L Smith and Quoc~V Le.
\newblock A bayesian perspective on generalization and stochastic gradient descent.
\newblock \emph{arXiv preprint arXiv:1710.06451}, 2017.

\bibitem[Sohl-Dickstein et~al.(2020)Sohl-Dickstein, Novak, Schoenholz, and Lee]{sohl2020infinite}
Jascha Sohl-Dickstein, Roman Novak, Samuel~S Schoenholz, and Jaehoon Lee.
\newblock On the infinite width limit of neural networks with a standard parameterization.
\newblock \emph{arXiv preprint arXiv:2001.07301}, 2020.

\bibitem[Su et~al.(2024)Su, Ahmed, Lu, Pan, Bo, and Liu]{su2024roformer}
Jianlin Su, Murtadha Ahmed, Yu~Lu, Shengfeng Pan, Wen Bo, and Yunfeng Liu.
\newblock Roformer: Enhanced transformer with rotary position embedding.
\newblock \emph{Neurocomputing}, 568:\penalty0 127063, 2024.

\bibitem[Touvron et~al.(2023)Touvron, Martin, Stone, Albert, Almahairi, Babaei, Bashlykov, Batra, Bhargava, Bhosale, et~al.]{touvron2023llama}
Hugo Touvron, Louis Martin, Kevin Stone, Peter Albert, Amjad Almahairi, Yasmine Babaei, Nikolay Bashlykov, Soumya Batra, Prajjwal Bhargava, Shruti Bhosale, et~al.
\newblock Llama 2: Open foundation and fine-tuned chat models.
\newblock \emph{arXiv preprint arXiv:2307.09288}, 2023.

\bibitem[Vaswani et~al.(2017)Vaswani, Shazeer, Parmar, Uszkoreit, Jones, Gomez, Kaiser, and Polosukhin]{vaswani2017attention}
Ashish Vaswani, Noam Shazeer, Niki Parmar, Jakob Uszkoreit, Llion Jones, Aidan~N Gomez, {\L}ukasz Kaiser, and Illia Polosukhin.
\newblock Attention is all you need.
\newblock \emph{Advances in neural information processing systems}, 30, 2017.

\bibitem[Vyas et~al.(2023)Vyas, Morwani, Zhao, Kaplun, Kakade, and Barak]{vyas2023beyond}
Nikhil Vyas, Depen Morwani, Rosie Zhao, Gal Kaplun, Sham Kakade, and Boaz Barak.
\newblock Beyond implicit bias: The insignificance of sgd noise in online learning.
\newblock \emph{arXiv preprint arXiv:2306.08590}, 2023.

\bibitem[Wang \& Aitchison(2024)Wang and Aitchison]{wang2024set}
Xi~Wang and Laurence Aitchison.
\newblock How to set adamw's weight decay as you scale model and dataset size.
\newblock \emph{arXiv preprint arXiv:2405.13698}, 2024.

\bibitem[Woodworth et~al.(2020)Woodworth, Gunasekar, Lee, Moroshko, Savarese, Golan, Soudry, and Srebro]{pmlr-v125-woodworth20a}
Blake Woodworth, Suriya Gunasekar, Jason~D. Lee, Edward Moroshko, Pedro Savarese, Itay Golan, Daniel Soudry, and Nathan Srebro.
\newblock Kernel and rich regimes in overparametrized models.
\newblock In Jacob Abernethy and Shivani Agarwal (eds.), \emph{Proceedings of Thirty Third Conference on Learning Theory}, volume 125 of \emph{Proceedings of Machine Learning Research}, pp.\  3635--3673. PMLR, 09--12 Jul 2020.
\newblock URL \url{https://proceedings.mlr.press/v125/woodworth20a.html}.

\bibitem[Wortsman et~al.(2023)Wortsman, Liu, Xiao, Everett, Alemi, Adlam, Co-Reyes, Gur, Kumar, Novak, et~al.]{wortsman2023small}
Mitchell Wortsman, Peter~J Liu, Lechao Xiao, Katie Everett, Alex Alemi, Ben Adlam, John~D Co-Reyes, Izzeddin Gur, Abhishek Kumar, Roman Novak, et~al.
\newblock Small-scale proxies for large-scale transformer training instabilities.
\newblock \emph{arXiv preprint arXiv:2309.14322}, 2023.

\bibitem[Xiao(2022)]{xiao2022eigenspace}
Lechao Xiao.
\newblock Eigenspace restructuring: a principle of space and frequency in neural networks.
\newblock In \emph{Conference on Learning Theory}, pp.\  4888--4944. PMLR, 2022.

\bibitem[Xiao et~al.(2019)Xiao, Pennington, and Schoenholz]{xiao2019disentangling}
Lechao Xiao, Jeffrey Pennington, and Sam Schoenholz.
\newblock Disentangling trainability and generalization in deep learning.
\newblock 2019.

\bibitem[Yang \& Hu(2021)Yang and Hu]{yang2021tensor}
Greg Yang and Edward~J Hu.
\newblock Tensor programs iv: Feature learning in infinite-width neural networks.
\newblock In \emph{International Conference on Machine Learning}, pp.\  11727--11737. PMLR, 2021.

\bibitem[Yang et~al.(2022)Yang, Hu, Babuschkin, Sidor, Liu, Farhi, Ryder, Pachocki, Chen, and Gao]{yang2022tensor}
Greg Yang, Edward~J Hu, Igor Babuschkin, Szymon Sidor, Xiaodong Liu, David Farhi, Nick Ryder, Jakub Pachocki, Weizhu Chen, and Jianfeng Gao.
\newblock Tensor programs v: Tuning large neural networks via zero-shot hyperparameter transfer.
\newblock \emph{arXiv preprint arXiv:2203.03466}, 2022.

\bibitem[Yang et~al.(2023)Yang, Yu, Zhu, and Hayou]{yang2023tensor}
Greg Yang, Dingli Yu, Chen Zhu, and Soufiane Hayou.
\newblock Tensor programs vi: Feature learning in infinite-depth neural networks.
\newblock \emph{arXiv preprint arXiv:2310.02244}, 2023.

\bibitem[Zhang et~al.(2021)Zhang, Bengio, Hardt, Recht, and Vinyals]{zhang2021understanding}
Chiyuan Zhang, Samy Bengio, Moritz Hardt, Benjamin Recht, and Oriol Vinyals.
\newblock Understanding deep learning (still) requires rethinking generalization.
\newblock \emph{Communications of the ACM}, 64\penalty0 (3):\penalty0 107--115, 2021.

\bibitem[Zhang et~al.(2018)Zhang, Wang, Xu, and Grosse]{zhang2018three}
Guodong Zhang, Chaoqi Wang, Bowen Xu, and Roger Grosse.
\newblock Three mechanisms of weight decay regularization.
\newblock \emph{arXiv preprint arXiv:1810.12281}, 2018.

\bibitem[Zhang et~al.(2024)Zhang, Peng, Zhao, Hu, Zhu, Zeng, and Hu]{zhang2024llasa}
Shuo Zhang, Boci Peng, Xinping Zhao, Boren Hu, Yun Zhu, Yanjia Zeng, and Xuming Hu.
\newblock Llasa: Large language and e-commerce shopping assistant.
\newblock \emph{arXiv preprint arXiv:2408.02006}, 2024.

\bibitem[Zhang et~al.(2023)Zhang, Roller, Goyal, Artetxe, Chen, Chen, Dewan, Diab, Li, Lin, et~al.]{zhang2023opt}
Susan Zhang, Stephen Roller, Naman Goyal, Mikel Artetxe, Moya Chen, Shuohui Chen, Christopher Dewan, Mona Diab, Xian Li, Xi~Victoria Lin, et~al.
\newblock Opt: Open pre-trained transformer language models, 2022.
\newblock \emph{URL https://arxiv. org/abs/2205.01068}, 3:\penalty0 19--0, 2023.

\bibitem[Zhao et~al.(2024)Zhao, Morwani, Brandfonbrener, Vyas, and Kakade]{zhao2024deconstructing}
Rosie Zhao, Depen Morwani, David Brandfonbrener, Nikhil Vyas, and Sham Kakade.
\newblock Deconstructing what makes a good optimizer for language models.
\newblock \emph{arXiv preprint arXiv:2407.07972}, 2024.

\end{thebibliography}
\bibliographystyle{tmlr}

\appendix
% \section{Appendix}

\end{document}